\documentclass{ieeeaccess}
\usepackage{cite}
\usepackage{amsmath,amssymb,amsfonts}
\usepackage{algorithmic}
\usepackage{graphicx}
\usepackage{textcomp}
\usepackage{listings}
\usepackage{array}
\usepackage{booktabs}
\usepackage{subfigure}
\usepackage[hyphens]{url}

\graphicspath{{figures/}}
\usepackage{bm}
\DeclareMathOperator*{\argmax}{arg\!\max}

\usepackage{xcolor}
\usepackage{pict2e}

\newsavebox{\ORCIDlogo}
\savebox{\ORCIDlogo}{%
\setlength{\unitlength}{\dimexpr 1em/256\relax}%
\begin{picture}(256,256)%
  \color[HTML]{A6CE39}\put(128,128){\circle*{256}}%
  \color{white}%
  \put(78.6,199.2){\circle*{20}}%
  \moveto(70.9,176,9)\lineto(86.3,176,9)\lineto(86.3,69.8)\lineto(70.9,69.8)%
  \closepath\fillpath%
  \moveto(108.9,176.9)\lineto(150.5,176.9)%
  \curveto(190.1,176.9)(207.5,148.6)(207.5 ,123.3)%
  \curveto(207.5,95,8)(186,69.7)(150.7,69.7)%
  \lineto(108.9,69.7)%
  \closepath\fillpath%
  \color[HTML]{A6CE39}%
  \moveto(124.3,83.6)\lineto(148.8,83.6)%
  \curveto(183.7,83.6)(191.7,110.1)(191.7,123.3)%
  \curveto(191.7,144.8)(178,163)(148,163)%
  \lineto(124.3,163)%
  \closepath\fillpath%
\end{picture}%
}

\newcommand\orcidicon[1]{\href{https://orcid.org/#1}{\usebox{\ORCIDlogo}}}

\usepackage{hyperref} 
\hypersetup{hidelinks} 

\makeatletter
\AtBeginDocument{\DeclareMathVersion{bold}
\SetSymbolFont{operators}{bold}{T1}{times}{b}{n}
\SetSymbolFont{NewLetters}{bold}{T1}{times}{b}{it}
\SetMathAlphabet{\mathrm}{bold}{T1}{times}{b}{n}
\SetMathAlphabet{\mathit}{bold}{T1}{times}{b}{it}
\SetMathAlphabet{\mathbf}{bold}{T1}{times}{b}{n}
\SetMathAlphabet{\mathtt}{bold}{OT1}{pcr}{b}{n}
\SetSymbolFont{symbols}{bold}{OMS}{cmsy}{b}{n}
\renewcommand\boldmath{\@nomath\boldmath\mathversion{bold}}}
\makeatother

\def\BibTeX{{\rm B\kern-.05em{\sc i\kern-.025em b}\kern-.08em
    T\kern-.1667em\lower.7ex\hbox{E}\kern-.125emX}}

\begin{document}
\history{Received 2 April 2026, accepted 18 April 2026, date of publication 23 April 2026} 
\doi{10.1109/ACCESS.2026.3687084} 

\title{ASVSim (AirSim for Surface Vehicles): A High-Fidelity Simulation Framework for Autonomous Surface Vehicle Research}
\author{\uppercase{Bavo Lesy*\orcidicon{0009-0000-9395-9437}}\authorrefmark{1},
\uppercase{Siemen Herremans*\orcidicon{0000-0001-7880-7144}}\authorrefmark{1},
\uppercase{Robin Kerstens\orcidicon{0000-0002-5127-4947}}\authorrefmark{2,3},
\uppercase{Jan Steckel\orcidicon{0000-0003-4489-466X}}\authorrefmark{2,3},
\uppercase{Walter Daems\orcidicon{0000-0001-6717-744X}}\authorrefmark{2,3}\IEEEmembership{Senior Member, IEEE},
\uppercase{Siegfried Mercelis}\orcidicon{0000-0001-9355-6566}\authorrefmark{1},
\uppercase{Ali Anwar\orcidicon{0000-0002-5523-0634}}\authorrefmark{1}\IEEEmembership{Member, IEEE}}

\address[1]{IDLab, Faculty of Applied Engineering, University of Antwerp - imec, 2000 Antwerp, Belgium}
\address[2]{Cosys-Lab, Faculty of Applied Engineering, University of Antwerp, 2020 Antwerp, Belgium}
\address[3]{Flanders Make Strategic Research Centre, 3920 Lommel, Belgium}
\tfootnote{This work was supported in part by the Horizon 2020 PIONEERS Project, funded by European Commission under Agreement 101037564; in part by the HORIZON  INNO2MARE project, funded by European Commission under Agreement 101087348; in part by the Research Foundation Flanders (FWO) under Grant 1SHAI24N.}

\markboth
{Lesy \headeretal: ASVSim: A High-Fidelity Simulation Framework for Autonomous Surface Vehicle Research}
{Lesy \headeretal: ASVSim: A High-Fidelity Simulation Framework for Autonomous Surface Vehicle Research}

\corresp{Corresponding author: Bavo Lesy (e-mail: bavo.lesy@uantwerpen.be).}

\begin{abstract}
The transport industry has recently shown significant interest in unmanned surface vehicles (USVs), specifically for port and inland waterway transport. These systems can improve operational efficiency and safety, which is especially relevant in the European Union, where initiatives such as the Green Deal are driving a shift towards increased use of inland waterways. At the same time, a shortage of qualified personnel is accelerating the adoption of autonomous solutions. However, there is a notable lack of open-source, high-fidelity simulation frameworks and datasets for developing and evaluating such solutions. To address these challenges, we introduce AirSim for Surface Vehicles (ASVSim), an open-source simulation framework specifically designed for autonomous shipping research in inland and port environments. The framework combines simulated vessel dynamics with marine sensor simulation capabilities, including radar and camera systems and supports the generation of synthetic datasets for training computer vision models and reinforcement learning (RL) agents. Built upon Cosys-AirSim, ASVSim provides a comprehensive platform for developing autonomous navigation algorithms and generating synthetic datasets. The simulator supports research of both traditional control methods and deep learning-based approaches. Through experiments in waterway segmentation and autonomous navigation, we demonstrate the capabilities of the simulator in these research areas. ASVSim is provided as an open-source project under the MIT license, making autonomous navigation research accessible to a larger part of the ocean engineering community. See ~\url{https://github.com/BavoLesy/ASVSim}.
\end{abstract}


\titlepgskip=-21pt

\maketitle

\section{Introduction}
\label{sec:introduction}
\PARstart{T}{here} is a significant research effort concerning autonomous navigation systems for USVs~\cite{gu2021autonomous,pietrzykowski2022autonomous}. However, developing and testing autonomous navigation systems in real-world environments is challenging due to high costs, safety risks, and the complex nature of maritime operations. For inland waterways, the complexity is further increased by narrow passages, locks, and dense traffic patterns and a frequent need to observe and avoid obstacles through the use of sensors. Infrastructure such as bridges, ports, and shore-based facilities may lead to sensor occlusion, creating blind spots that autonomous systems must account for. 
Additionally, varying wind and current conditions can significantly impact a vessel, requiring robust control strategies to maintain safe navigation. Because of these challenges, simulation has emerged as a crucial tool for developing and validating navigation algorithms, in a safe and cost-effective manner~\cite{bookFossen, zhang2024path, SERIGSTAD20181}. For these purposes, it is important that the simulation closely represents the real world to reduce the effort needed to deploy simulated solutions on real vessels. However, a current gap exists in the availability of shipping simulators for autonomy research, with most existing simulators either being designed for human operator training, being proprietary, or lacking the visual realism (or fidelity) needed for accurate camera and sensor simulation. More recently, there is an increasing interest in the potential of deep learning methods in a shipping context, such as obstacle detection~\cite{jmse12010190} or Reinforcement Learning (RL) for navigation~\cite{zhou2019rl}. RL has emerged as a promising approach for USVs, partly because of its ability to generalize across different vessel types and operating conditions~\cite{lesyRobustOfflineReinforcement2025a}. However, the USV domain currently lacks accessible, high-fidelity simulation software that can accurately model vessel dynamics and generate realistic sensor data for development of autonomous systems that perceive their environment and navigate within it.

The main contribution of this paper is ASVSim\footnote{https://github.com/BavoLesy/ASVSim}, an open-source, high-fidelity simulator specifically designed for autonomous shipping research. In the context of simulation, we use the term high-fidelity to refer to three complementary dimensions. First, photorealistic visual rendering enables the generation of realistic synthetic data for perception research (Sections~\ref{sec:dataset} and~\ref{sec:cv}). Second, physics-based vessel dynamics use the Fossen maneuvering model to accurately reproduce vessel motion (Section~\ref{sec:dynsim}). Third, sensor simulation provides camera rendering and a radar emulation that approximates marine radar imagery from LiDAR point clouds using a point spread function (PSF) (Section~\ref{sec:radsim}). The most notable modifications and extensions to the underlying Cosys-AirSim~\cite{cosysairsim2023jansen} framework are summarized in Fig.~\ref{fig:sim_overview}. ASVSim is tailored for Inland Waterway Transport (IWT), where the focus is more on the navigation of vessels through narrow passages and locks, and less on the dynamics of the vessel itself. The simulator currently provides 3-degree-of-freedom (3DoF) maneuvering dynamics, sensor simulations, and environmental conditions such as wind and currents. This enables researchers to develop, train, and test autonomous navigation algorithms in a configurable setting. As will be demonstrated throughout this work, ASVSim allows for the creation of high-fidelity environments, essential for image-based sensor research. Because of this visual fidelity, ASVSim can be used to generate realistic synthetic datasets to train computer vision models. We demonstrate how the simulator can be used for training and validation of autonomy solutions, using examples such as dataset collection and waterway segmentation. Autonomous navigation tasks are demonstrated as use-cases of the simulator, including autonomous channel navigation and the training of an RL agent to navigate a vessel to a goal whilst avoiding obstacles.

\Figure[t!](topskip=0pt, botskip=0pt, midskip=0pt)[width=0.48\textwidth]{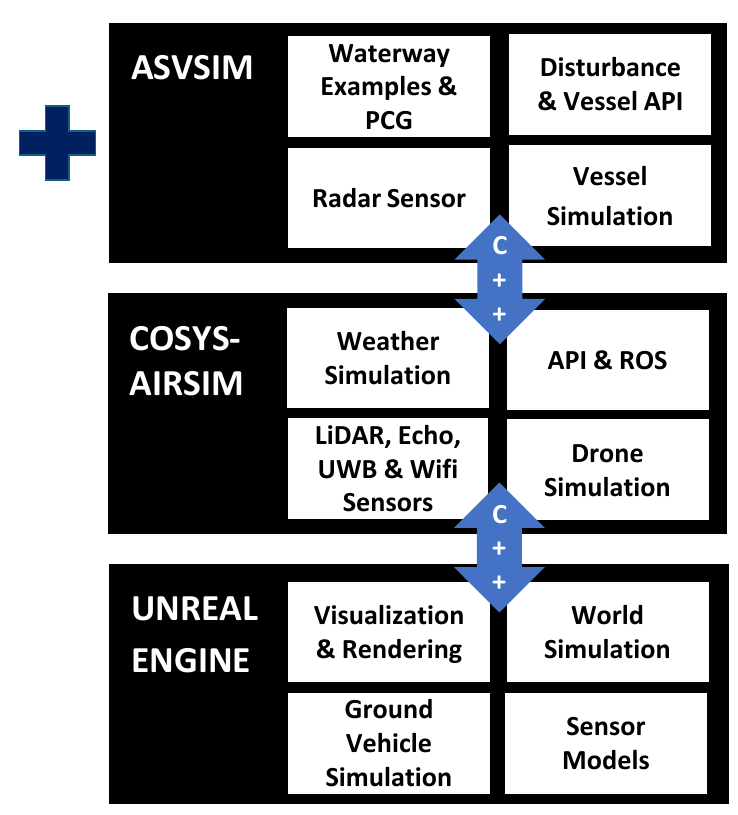}
{\textbf{Overview of the ASVSim framework, showing the extensions and modifications to Cosys-AirSim. The simulator integrates specialized maritime dynamics models, radar simulation, and other features for autonomous vessel research.}\label{fig:sim_overview}}

The paper is structured as follows: in Section~\ref{sec:related_works}, we review related works in the field of shipping simulators. In Section~\ref{sec:dynsim} and~\ref{sec:radsim}, we describe the mathematical models underlying ASVSim, covering vessel dynamics and radar simulation. In Section~\ref{sec:airsim}, we present the ASVSim framework in detail, including its dataset generation capabilities, procedural environment generation, and the Python and Matlab API for vessel control. In Section~\ref{sec:experiments}, we present the results of our autonomy experiments, covering waterway segmentation and path planning. In Section~\ref{sec:future_work}, we discuss the results and outline future research directions. Finally, in Section~\ref{sec:conclusion}, we provide a conclusion.

\section{Related Works}\label{sec:related_works}
\begin{table*}[htbp]
    \centering
    \caption{\textbf{Comparison of surface vessel simulators. Visual: camera-based perception sensors. Ranging: LiDAR and/or radar sensors.}}
    \label{tab:simulator_comparison}
    \renewcommand{\arraystretch}{1.2}
    \setlength{\tabcolsep}{4pt}
    \begin{tabular}{|l|c|c|c|c|c|c|l|}
        \hline
        \textbf{Simulator} & \textbf{Engine} & \textbf{Visual Fidelity} & \textbf{Dynamics} & \textbf{Open-Source} & \textbf{Visual} & \textbf{Ranging} & \textbf{Primary Usage} \\
        \hline
        Kongsberg K-Sim~\cite{kongsberg2025k-sim} & Proprietary & 3D & Proprietary & $\times$ & $\times$ & $\times$ & Maritime training \\
        \hline
        Nautis~\cite{nautis2025} & Proprietary & 3D (Photorealistic) & Proprietary & $\times$ & $\times$ & $\times$ & Maritime training \\
        \hline
        W\"{a}rtsil\"{a} R\&D Sim.~\cite{wartsila2025rdsim} & Proprietary & 3D & Proprietary & $\times$ & $\times$ & $\times$ & Autonomy research \\
        \hline
        SIMTECH ASHSS~\cite{simtech2025} & Proprietary & 3D (Photorealistic) & Proprietary & $\times$ & $\times$ & $\times$ & Autonomy research \\
        \hline
        AILiveSim~\cite{leudet2019ailivesim} & UE & Photorealistic & Proprietary & $\times$ & \checkmark & \checkmark & Autonomy research \\
        \hline
        MSS~\cite{fossen2004mss} & Matlab & None & Fossen & \checkmark & $\times$ & $\times$ & Control systems \\
        \hline
        uSimMarine~\cite{moosivp} & C++ & 2D & Simple & \checkmark & $\times$ & $\times$ & Autonomy research \\
        \hline
        VRX~\cite{bingham2019vrx} & Gazebo & 3D & Simple & \checkmark & \checkmark & \checkmark & Autonomy research \\
        \hline
        \textbf{ASVSim (ours)} & UE5 & 3D (Photorealistic) & Fossen & \checkmark & \checkmark & \checkmark & Autonomy research \\
        \hline
    \end{tabular}
\end{table*}

Mission simulation has long been an essential tool in the shipping industry, primarily for training and certification of maritime personnel. The evolution of these simulators has been driven by the need for increasingly realistic training environments that can prepare personnel for operating modern vessels. Industry-standard simulators such as Kongsberg K-Sim~\cite{kongsberg2025k-sim} and Nautis~\cite{nautis2025} have become more and more present in maritime education, providing highly realistic environments with accurate vessel dynamics, environmental conditions, and operational scenarios. These commercial systems excel in their primary use case: training maritime professionals in vessel handling, navigation, and emergency procedures across various vessel types and maritime conditions. These simulators offer realistic physics models and visualization of maritime scenarios with detailed rendering of weather conditions and accurate representations of navigational sensors and port infrastructure. Many include replicas of actual bridge equipment and instrumentation, closely mirroring real-world vessel operations. However, as these systems were designed primarily for human operators, they are not directly suitable for autonomy research. Additionally, these simulators are mostly closed-source and licenses can be expensive. 

The limitations of these traditional training simulators have become apparent as the maritime industry increasingly embraces automation. While platforms such as SIMTECH~\cite{simtech2025} and the Wärtsilä R\&D Simulator~\cite{wartsila2025rdsim} have begun incorporating features for autonomous vessel development, their proprietary nature and commercial focus create substantial barriers for academic research and open innovation. Entertainment-oriented simulators such as European Ship Simulator~\cite{europeanshipsim}, though more accessible, lack the physical accuracy and sensor simulation capabilities essential for meaningful autonomy research. AILiveSim~\cite{leudet2019ailivesim}, built on Unreal Engine, provides camera, LiDAR, and radar sensor simulation for autonomous vessel development, but is not open-source. Several open-source maritime simulators have emerged to address the research gap in autonomous navigation, but each has significant limitations. MOOS-IVP (uSimMarine)~\cite{moosivp}, a widely used open-source platform, provides a lightweight framework for autonomous vessel control with a publish-subscribe communication architecture that makes it straightforward to integrate new autonomy modules. However, it relies on simplified vessel dynamics, 2D top-down visualization (via pMarineViewer), and lacks native ranging sensor simulation (e.g., radar or LiDAR). While data can in principle be logged through the MOOS publish-subscribe system, the 2D visualization and absence of 3D sensor modalities make it unsuitable for generating realistic synthetic perception datasets. These properties make uSimMarine well suited for rapid prototyping and lightweight control experiments where visual realism is not required, but limit its applicability to perception-driven and learning-based autonomy research. Similarly, the Marine Systems Simulator (MSS)~\cite{fossen2004mss}, a Matlab/Simulink-based library for marine control systems development, offers elaborate physics models for various vessel types (ships, underwater vehicles, and unmanned underwater vehicles (UUVs)), but lacks integrated sensor simulation and comprehensive visualization capabilities. These limitations significantly constrain the effectiveness of both simulators for developing and training autonomous maritime navigation systems. VRX~\cite{bingham2019vrx}, the Virtual RobotX simulator built on Gazebo, provides an open-source platform for USV autonomy research with ROS integration, 3DoF dynamics with wave and wind plugins, and camera and LiDAR sensors, but lacks radar simulation and photorealistic rendering. Simultaneously, the field of autonomous driving has benefited greatly from several open-source simulators with high visual fidelity such as CARLA~\cite{dosovitskiy2017carla} and Cosys-AirSim~\cite{cosysairsim2023jansen}. These simulators excel at replicating real-world conditions through detailed graphics, precise physics models, and advanced sensor simulation. Such high-fidelity environments allow algorithms that were developed in simulation to transfer more effectively to real-world applications~\cite{bu2021carla}. Another benefit of this fidelity is the ability and flexibility to effectively recreate real-world scenarios. Simulators such as Stonefish~\cite{stonefish}, OceanSim~\cite{song2025oceansim} and NauSim~\cite{nausim}, offer realistic underwater environments and diverse sensor capabilities for UUV development and testing. Additionally, UNav-Sim~\cite{unavsim} and HoloOcean~\cite{holoocean} leverage the advanced capabilities of Unreal Engine~\cite{unreal2025} to further enhance simulation realism. However, these underwater-focused simulators are not suitable for inland shipping applications, as they primarily model underwater physics and UUV models and dynamics, in addition to the absence of dynamic obstacles and inland waterway environments. 

Therefore, as summarized in Table~\ref{tab:simulator_comparison}, a critical gap exists in the inland shipping domain particularly for USVs: the lack of open-source, high-fidelity simulators that have the physical realism and the functionality required for inland autonomy research. Notably, this shortage is also present in high-quality datasets, with maritime data typically remaining limited, fragmented, or proprietary. For an overview of marine datasets, the reader is referred to~\cite{su2023survey} and~\cite{trinh2024survey}. ASVSim addresses these challenges by providing an open-source simulator with high visual fidelity, serving the needs of autonomous navigation research, with particular attention to inland waterways and ports. Our platform prioritizes ease of use for deep learning and robotics research applications. Furthermore, the simulator can generate comprehensive, diverse datasets essential for training and validating autonomy solutions.

\section{Dynamics Simulation}\label{sec:dynsim}
\Figure[t!](topskip=0pt, botskip=0pt, midskip=0pt)[width=0.48\textwidth]{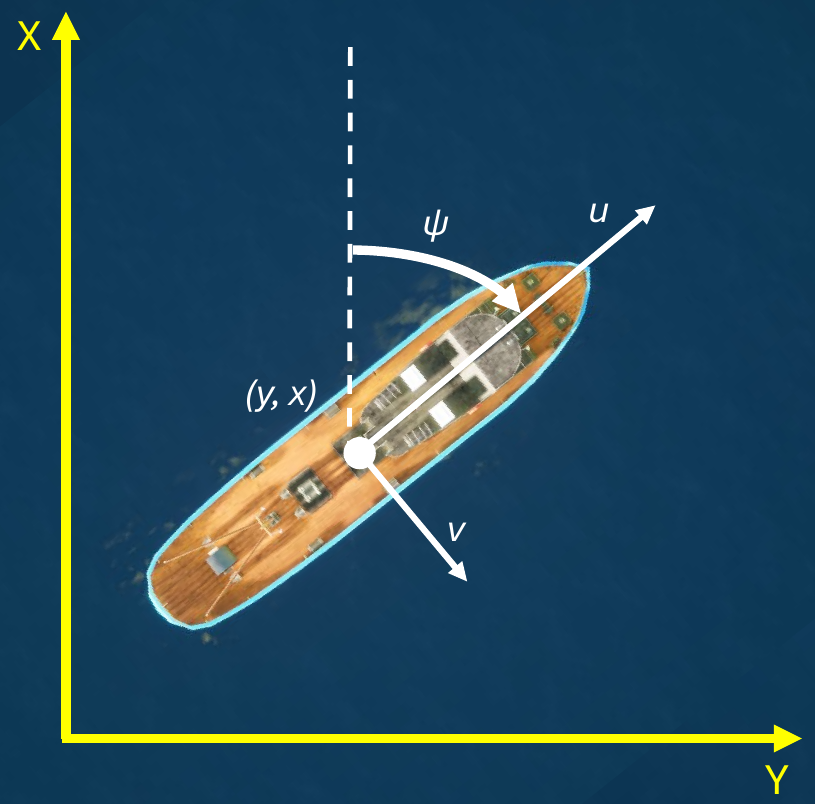}
{\textbf{Visualization of the Fossen vessel model used in the simulation. The vessel has three degrees of freedom: surge, sway, and yaw.}\label{fig:3dof}}

Since the main focus of this paper is inland waterway navigation, we opt for a 3DoF model, ignoring the heave, roll and pitch of the vessel. The 3DoF maneuvering model is the standard formulation for inland vessels, where sheltered, calm water conditions make vertical dynamics negligible~\cite{bookFossen}. As a result, wave excitation forces, vessel responses in the vertical plane, and effects such as propeller-hull interactions are not captured. Extending the simulator to 6DoF is architecturally straightforward, but was not pursued because 6DoF models introduce substantially more hydrodynamic parameters, making the system underdetermined without extensive experimental data for identification. As a result, published 6DoF parameter sets for inland vessel types are rarely available in the literature. Researchers requiring 6DoF dynamics can extend the model by providing the additional mass, damping, and Coriolis matrices for heave, roll, and pitch. The impact on sim-to-real transfer is limited, as the same Fossen 3DoF model is widely used for real-world maritime control system design~\cite{fossen2004mss}. Specifically, we implement the 3DoF maneuvering model (see Fig.~\ref{fig:3dof}), as presented by~\cite{bookFossen}. This model allows us to express the surge, sway and rotational velocity of the vessel, based on the force inputs and hydrodynamical constants that define the mass, damping and Coriolis and centripetal effects. Throughout this work, we will be using bold notation for vectors (e.g. $\boldsymbol{x}$) and the dot notation for the derivatives of these vectors (e.g. $\boldsymbol{\dot{x}}$). Additionally, matrices are denoted as capitalized bold symbols (e.g. $\boldsymbol{M}$).

\subsection{Kinematics}
We describe the vessel's position with its positional coordinates and heading (in the Earth-fixed frame), $\boldsymbol{\eta} = [x, y, \psi]$. Secondly, we include the surge, sway and rotational velocity using $\boldsymbol{\nu} = [u, v, r]$ in the body-fixed frame. The kinematics of these vectors are described by~\eqref{eqn:kinematics}. 

\begin{align}
\begin{split}
\label{eqn:kinematics}
    \boldsymbol{\dot{\eta}} &= \boldsymbol{R_{rot}(\psi)\nu} \\
    \boldsymbol{R_{rot}(\psi)} &= 
    \begin{bmatrix} 
    cos(\psi) & -sin(\psi) & 0 \\ 
    sin(\psi) &  cos(\psi) & 0 \\
    0 & 0 & 1
    \end{bmatrix}
\end{split}
\end{align}

Furthermore, we allow for currents, which are described using a velocity vector $\boldsymbol{\dot{\eta}_c}$ in the Earth-fixed frame. These can be translated to a local-frame current velocity vector $\boldsymbol{\nu_c}$ via the (orthonormal) rotation matrix:

\begin{align}
    \boldsymbol{\nu_c} &= \boldsymbol{R_{rot}(\psi)^T \dot{\eta}_c}\\ \label{eqn:define_current}
    \boldsymbol{\dot{\eta}_c} &= \left[ V_c\cdot cos(\beta_c), V_c\cdot sin(\beta_c), 0 \right]^T
\end{align}

\noindent where $V_c$ describes the velocity of the current (in $m/s$) and $\beta_c$ denotes the heading of the current in the Earth-fixed frame (in radians). Finally, the relative velocity vector $\boldsymbol{\nu_r}$ is defined as the difference between the velocity of the ego vessel and the velocity of the current.

\begin{equation}
    \boldsymbol{\nu_r} = \boldsymbol{\nu} - \boldsymbol{\nu_c}
    \label{eqn:relative_velocity}
\end{equation}

\subsection{Dynamics}
\label{subsection:dynamics}

The dynamics are specified by five vessel-dependent $3\times3$ matrices. $\boldsymbol{M_{RB}}$ and $\boldsymbol{M_A}$ are the rigid-body and hydrodynamical mass matrices, defining the (added) mass of the vessel, the centers of gravity and the inertia around the yaw axis. $\boldsymbol{C_{RB}(\nu)}$ and $\boldsymbol{C_{A}(\nu_r)}$ define the rigid body and hydrodynamical Coriolis (and centripetal) matrices. Finally, $\boldsymbol{D(\nu_r)}$ represents the damping force acting on the vessel. Importantly, $\boldsymbol{D(\nu_r)}$ is allowed to contain both linear and non-linear damping elements, such as $u^2, v^2$ and $r^2$. Equation~\eqref{eqn:dynamics} fully describes the dynamics, subject to an external control vector $\boldsymbol{\tau}$ and wind $\boldsymbol{\tau_{wind}}$. 

\begin{align}
\label{eqn:dynamics}
\begin{split}
    \boldsymbol{M_{RB}\dot{\nu} + C_{RB}(\nu)\nu + M_{A}\dot{\nu_r} +} \\ \boldsymbol{C_A(\nu_r)\nu_r + D(\nu_r)\nu_r = \tau + \tau_{wind}}
\end{split}
\end{align}

Many practical models ignore currents (i.e. $\boldsymbol{\nu_c=0}$, where $\boldsymbol{0}$ denotes a vector containing only zeros)~\cite{zheng2014trajectory, kristiansen2024modelling}, allowing a simplification of the dynamics by defining $\boldsymbol{M} = \boldsymbol{M_{RB} + M_A}$ and $\boldsymbol{C} = \boldsymbol{C_{RB} + C_A}$. This simplified model (ignoring currents) is described in~\eqref{eqn:dynamics_no_current}.

\begin{equation}
\label{eqn:dynamics_no_current}
    \boldsymbol{M\dot{\nu} + C(\nu)\nu + D(\nu)\nu = \tau + \tau_{wind}}
\end{equation}

As we believe that currents and wind are important factors to consider in autonomous navigation research, we include these currents, under the assumption that they are irrotational and (approximately) constant. These assumptions are commonly made when modeling vessels, since they allow for practical parameter estimation on real-world vessels. More details on this are provided in Section~\ref{sec:currents}.

\subsection{Control Forces}
The simulator allows to place an arbitrary amount of thrusters. Each of them is controlled by setting a thrust force $F_i$ (in Newton) and the desired angle $\theta_i$ (where $i$ is the index of the thruster). Additionally, the position of each thruster can be configured using $d_{x_i}$ and $d_{y_i}$, which represent the position of the thruster with regard to the center of gravity (COG). For example, a stern thruster can be created by placing it at the back of the vessel and limiting the minimal and maximal turning angle. Similarly, a bow thruster can be created by adding a thruster at the front of the vessel, and fixing its angle. We model the force of the thrusters on the vessel as follows:

\begin{equation}
\label{eqn:thruster}
    \boldsymbol{\tau} = 
    \resizebox{0.75\columnwidth}{!}{$\begin{bmatrix} 
    \cos(\theta_1)       & \ldots & \cos(\theta_n)\\ 
    \sin(\theta_1)        & \ldots & \sin(\theta_n) \\
    d_{x_1}\sin(\theta_1) - d_{y_1}\cos(\theta_1) &  \ldots & d_{x_n}\sin(\theta_n) - d_{y_n}\cos(\theta_n)
    \end{bmatrix}$} \times \boldsymbol{f}
\end{equation}

\noindent where $\boldsymbol{f}=[F_1, F_2, \ldots, F_n]^T$ represents the vector containing all thruster forces. Similarly, $\boldsymbol{\theta}=[\theta_1, \theta_2, \ldots, \theta_n]$ represents all thruster angles (in the local frame). The control input vector $\boldsymbol{\tau}$ contains the surge, sway and yaw components that are added to the system. 

\subsection{Currents}
\label{sec:currents}

\Figure[t!](topskip=0pt, botskip=0pt, midskip=0pt)[width=1.0\textwidth]{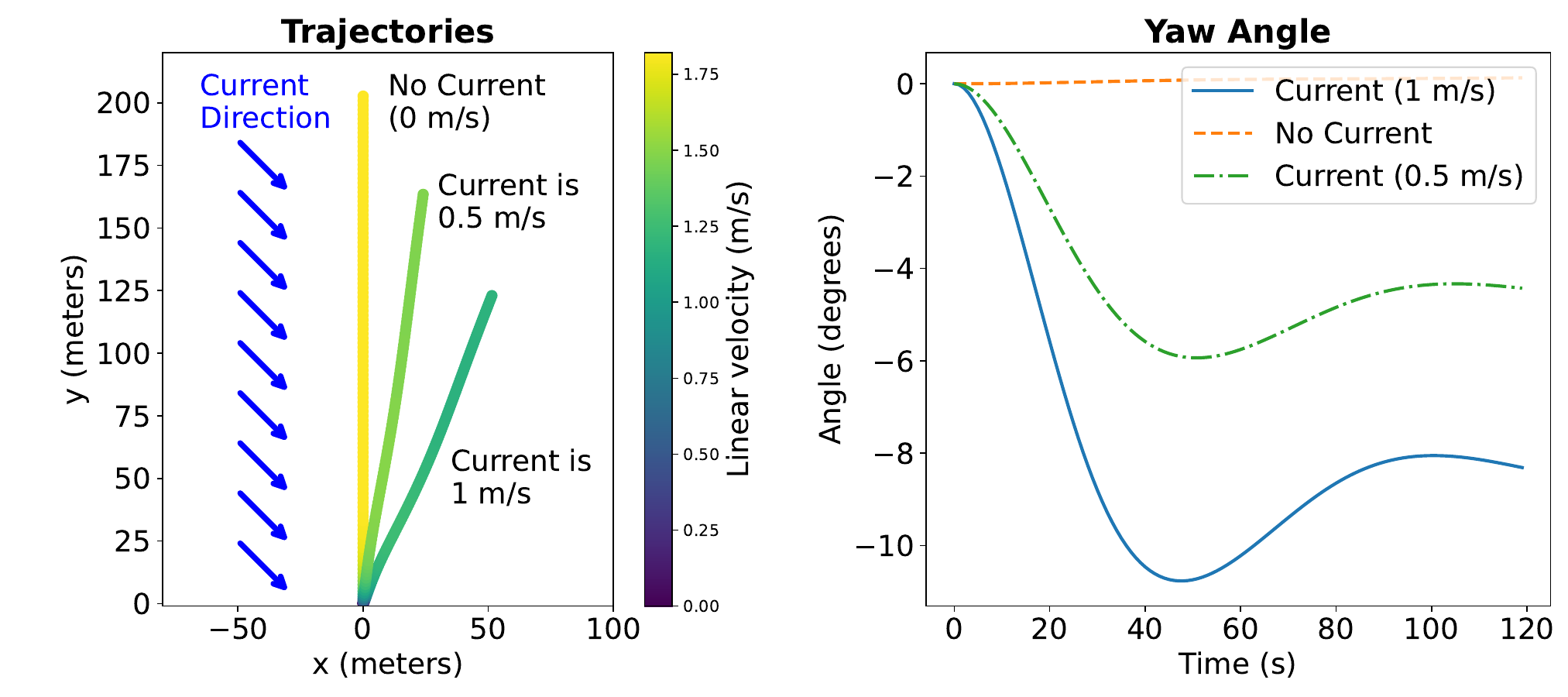}
{\textbf{Sailing the MilliAmpere Ferry with thrust force at half ahead (250 N) with a straight thruster. Currents are simulated with an impact angle of $135^{\circ}$ (top left). Left image: the recorded trajectories, starting from $x=0$, $y=0$. Right image: the rotation around the yaw axis, induced by the current.}\label{fig:trajectory}}

Equation~\eqref{eqn:dynamics} already includes the effect of currents, since it is parameterized on both $\boldsymbol{\nu}$ and $\boldsymbol{\nu_r}$. However, separate parameter estimation for the rigid-body matrices (such as $\boldsymbol{M_{RB}}$) and the hydrodynamical matrices (such as $\boldsymbol{M_A}$) is impractical and uncommon in research, since it makes real-world parameter estimation impractical. Therefore, we follow the common assumption that currents are irrotational and constant over short time spans~\cite{YOON20032379, bookFossen, RobustLinearMSS}. The irrotational property refers to the absence of a sway component in the current, as denoted in~\eqref{eqn:define_current}. The constant property assumes that there is negligible acceleration of the current in the Earth-fixed frame, i.e., $\boldsymbol{\ddot{\eta}_c} \approx \boldsymbol{0}$. Under these ocean current assumptions, it is shown that, when we parametrize the $\boldsymbol{C_{RB}}$ matrix independent of surge and sway, the dynamics can be described as follows:

\begin{align}
    \boldsymbol{M\dot{\nu_r} + C(\nu_r)\nu_r+D(\nu_r)\nu_r = \tau+\tau_{wind}},
\end{align}

where $\boldsymbol{M}$ and $\boldsymbol{C}$ are defined as in Section~\ref{subsection:dynamics}. This is a practical model, since it only requires parameter estimation for a singular mass, Coriolis and damping matrix. It is trivial to compute the absolute velocity from the relative velocity using~\eqref{eqn:relative_velocity}.

\subsection{Implemented Models}
ASVSim allows to easily add specific damping and Coriolis models to compute $\boldsymbol{C(\nu_r)}$ and $\boldsymbol{D(\nu_r)}$. Four models are already implemented and available in the simulator:
\begin{itemize}
    \item MilliAmpere Ferry, an autonomous passenger ferry at the Norwegian University of Science and Technology (NTNU)~\cite{pedersen2019optimization},
    \item Qiuxin No.5, a small research vessel at the Wuhan University of Technology~\cite{you2023novel},
    \item CyberShip II, a 1:70 scale replica of a supply ship, created for towing tank experiments~\cite{SKJETNE2004203}.
    \item A full scale Mariner class cargo vessel~\cite{chislett1965planar}.
\end{itemize}

For the full scale vessel, we also provide a generic method to include shallow-water effects. We base ourselves on the (corrected) model by D. Clarke (1997) ~\cite{CLARKE1997117}, appropriate when the draft of the vessel is at most 80\% of the waterway depth. As an illustrative example of the simulated effect of currents on a vessel, we include Fig.~\ref{fig:trajectory}. This figure shows the recorded trajectories of the MilliAmpere ferry under no, low and moderate current velocities. As expected, currents with a higher velocity make the ferry move more from a straight path, and introduces torque on the vessel. Additionally, the current in the example decreases the speed of the vessel, as its impact angle has a component that is opposite to the thrust force direction.

\section{Radar Simulation}\label{sec:radsim}
Marine radar units are still dominant when it comes to localization, navigation, and collision avoidance in maritime environments. Although cameras and LiDAR systems are increasing in popularity, they cannot compete with the range that can be achieved with radar~\cite{siciliano2008springer}. Also, because of the familiarity between radar and the shipping industry, it makes sense to include a radar simulation in this simulation-based research. Cosys-Airsim~\cite{cosysairsim2023jansen} has an accurate radar simulation based on ray-tracing techniques; however, this is based on a frequency modulated continuous wave (FMCW) radar unit originally introduced by~\cite{schouten2021simulation}, which is inherently different from the single-frequency pulse-echo systems commonly used in maritime and inland settings. For this reason, this research mimics a marine radar by using a modified version of the original AirSim~\cite{shah2018airsim} LiDAR implementation. The simulated LiDAR provides, at a predetermined frame rate that can be set similar to the rotation speed of the radar unit, a dense 3D point cloud of the surroundings covering a field-of-view of 360 degrees azimuth and a predetermined elevation. To mimic the performance of a maritime radar such as the Furuno DRS12A, as described in~\cite{furuno2025}, the elevation was set to 20 degrees with a rational angle of 36 RPM. This unit has a range of 69 NM. In this scenario, a maximum range of 5 km (2.7 NM) was chosen. This range is sufficient because the depth of measurement rarely exceeds this value due to occlusion by the buildings and infrastructure present in inland waterways. Furthermore, objects located outside of this 5 km range are deemed not relevant in this use-case and are in turn omitted to limit computing requirements. This simulated approach also offers the added benefit that the dense LiDAR point cloud contains the maximum number of reflections, which allows us to generate datasets with varying levels of sparsity, mimicking scenarios with reduced vision.

\Figure[t!](topskip=0pt, botskip=0pt, midskip=0pt)[width=1.0\textwidth]{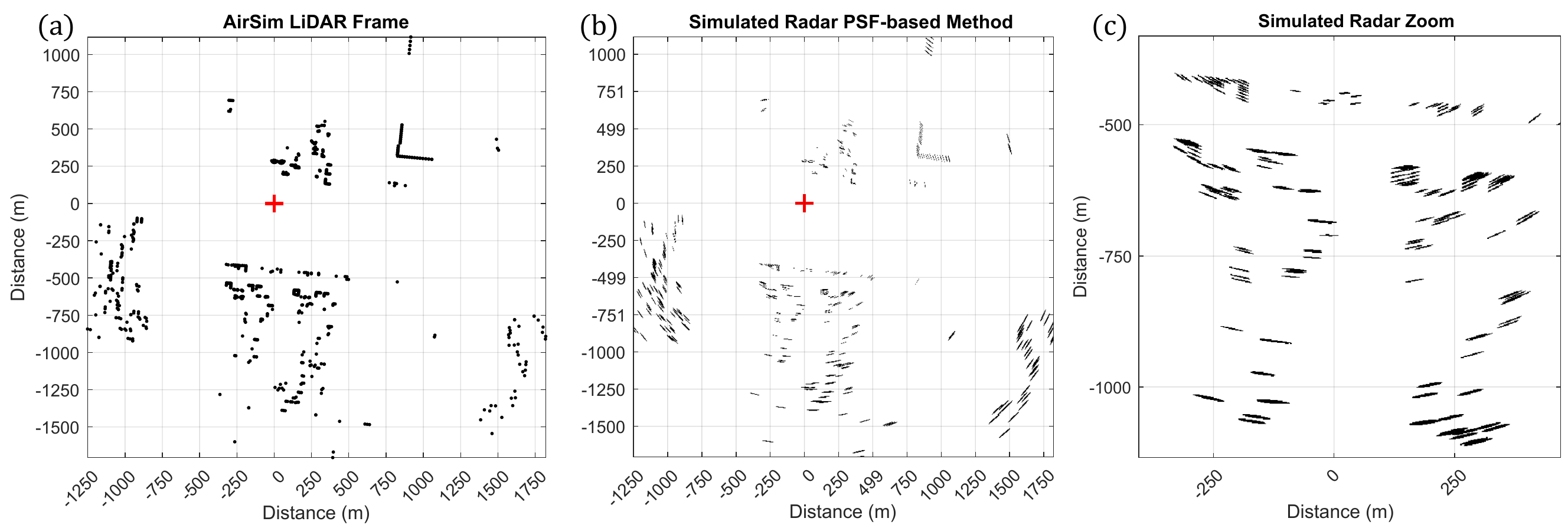}
{\textbf{(a) The raw LiDAR point cloud obtained in the simulator. (b) The emulated radar data obtained using the PSF-based method (c) A subset of the center image, displaying the PSF varying over range and angle.}\label{fig:radarDataApprox}}

\subsection{Radar Detection Approximation and Visualization}
Fig.~\ref{fig:radarDataApprox} illustrates the process used to mimic maritime radar data, starting from the LiDAR sensor. An example LiDAR frame is shown in Fig.~\ref{fig:radarDataApprox}a. Each simulated reflection obtained by the light pulses results in a set of cartesian coordinates. To mimic the typical visualization found on maritime radar displays, these LiDAR detection coordinates should be rasterized into an image, and should be convolved with the scanning radar PSF that can be expected at the locations of these points, as illustrated in Fig.~\ref{fig:radarDataApprox}b. Fig.~\ref{fig:radarDataApprox}c shows a close up of the same scene, highlighting the shape and orientation of the radar PSFs. The shape and orientation of this PSF is dependent on the properties of the radar, the range, and angle with respect to the location of the radar.

Starting from a set of input reflector points $\mathbf{p}$, obtained from the LiDAR simulation, the locations of the nearest and furthest points, $\mathbf{p}_{\min}$ and $\mathbf{p}_{\max}$, with range $\Delta \mathbf{p} = \mathbf{p}_{\max}- \mathbf{p}_{\min}$, and image size $G$, the normalized point coordinates in the new image for each point $\mathbf{p}'_i$ are:
\begin{equation}
    \mathbf{p}'_i = \left\lfloor \left( \frac{\mathbf{p}_i - \mathbf{p}_{\min}}{\Delta \mathbf{p}} \right) \cdot G  \right\rfloor 
\end{equation}
The direction vector $\mathbf{d}_i$ from radar position $\boldsymbol{p_r}$ to normalized point $\mathbf{p}'_i$ is:
\begin{equation}
    \mathbf{d}_i = \mathbf{p}'_i - \boldsymbol{p_r} 
\end{equation}
The range $r_i$ is the Euclidean norm of $\mathbf{d}_i$:
\begin{equation}
    r_i = \|\mathbf{d}_i\|_2
\end{equation} 
As the radar PSF has its rotation with respect to the (centralized) location of the radar, the angle $\theta_i$ can be found as:
\begin{equation}
    \theta_i = \arctan2(d_{i,y}, d_{i,x}) 
\end{equation}
where $d_{i,x}$ and $d_{i,y}$ are the $x$ and $y$ components of $\mathbf{d}_i$, respectively. The basic shape of the radar PSF will be depicted as an ellipsoid~\cite{richards2010principles}, of which the semi-major axis $a_i$ and semi-minor axis $b_i$ are:
\begin{align}
    a_i &= r_i \tan\left(\frac{\alpha}{2}\right) \frac{G}{\Delta p_x} \\
    b_i &= r_i \tan\left(\frac{\beta}{2}\right) \frac{G}{\Delta p_y}
\end{align}
where $\alpha$ is the horizontal beam width of the radar unit in radians, $\beta$ a scalar which is manually set to obtain the correct range resolution. $\Delta p_x$ and $\Delta p_y$ are the $x$ and $y$ components of $\Delta \mathbf{p}$, respectively.
To fill-in the correct pixels in the new image, it is not possible to use a standard convolution method as our PSF varies for every different point in the image. Therefore, we use the ellipse indicator function $E_{i}(x, y)$ is:
\begin{equation}
    E_{i}(x, y) = \begin{cases} 1 & \text{if } \frac{x_r^2}{a_i^2} + \frac{y_r^2}{b_i^2} \le 1 \\ 0 & \text{otherwise} \end{cases}
\end{equation}
where:
\begin{align}
    x_c &= x - p'_{i,x} \\
    y_c &= y - p'_{i,y} \\
    x_r &= x_c \cos(\theta_i) + y_c \sin(\theta_i) \\
    y_r &= -x_c \sin(\theta_i) + y_c \cos(\theta_i)
\end{align}
and $p'_{i,x}$ and $p'_{i,y}$ are the $x$ and $y$ components of $\mathbf{p}'_i$, respectively.
The radar PSF $P(x, y)$ is then the binary image that is obtained when taking the maximum of all ellipse indicator functions:
\begin{equation}
    P(x, y) = \max_{i} E_{i}(x, y)
\end{equation}

\subsection{Scope and Limitations}
The PSF-based radar approximation faithfully emulates the geometric characteristics of a pulse-echo marine radar: the ellipsoidal resolution cell shape follows directly from the radar's azimuth beam width and range resolution~\cite{richards2010principles}, and the spatially-varying PSF correctly models how resolution degrades with range. Shadowing of objects behind solid structures is inherently captured by the underlying ray-casting engine.

However, several phenomena present in real marine radar are not modeled. Sea clutter, the backscatter from the water surface, is a noise source in marine radar typically modeled using K-distribution or Rayleigh statistics~\cite{ward2013sea}. Our simulation does not generate sea surface returns, meaning the radar image appears cleaner than real-world recordings. We note that sea clutter is primarily a concern in open-ocean conditions and is less prevalent in sheltered inland waterways, which are the focus of this work. Multipath reflections, sidelobe returns, and radar cross-section (RCS) variations between different materials are also not captured, as the LiDAR-based approach treats all surfaces as ideal reflectors. These limitations make the current radar simulation most suitable for evaluating detection and navigation algorithms in terms of geometric accuracy, while quantitative radar signal fidelity requires validation against real marine radar recordings, which we identify as future work.

\section{AirSim for Surface Vehicles}\label{sec:airsim}
\Figure[t!](topskip=0pt, botskip=0pt, midskip=0pt)[width=0.48\textwidth]{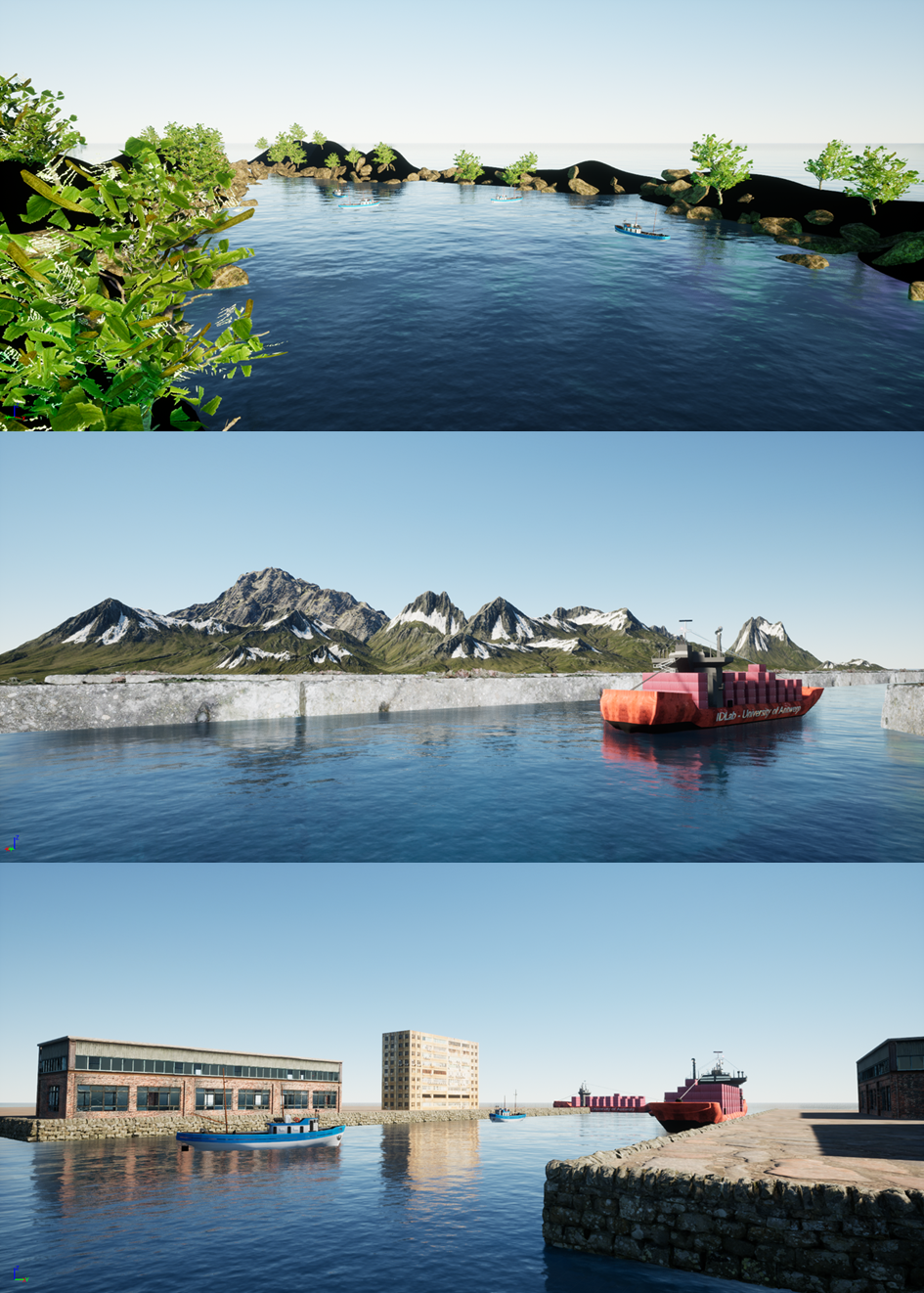}
{\textbf{Overview of the example environments currently included in ASVSim.}\label{fig:sim_envs}}
\Figure[t!](topskip=0pt, botskip=0pt, midskip=0pt)[width=1.0\textwidth]{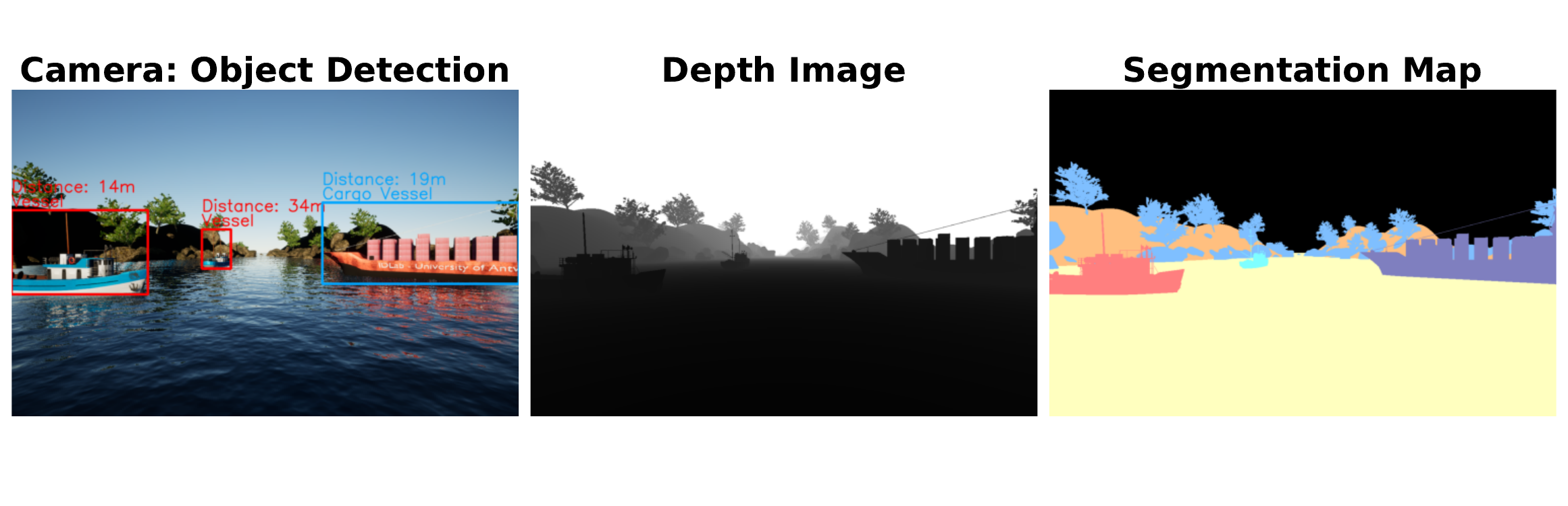}
{\textbf{Example output from generated dataset. From left to right: bounding boxes generated automatically for vessels in the environment, depth camera measurements in grayscale (0 to 255m), segmentation map.}\label{fig:cv_vessels}}

ASVSim is an extension of Cosys-AirSim~\cite{cosysairsim2023jansen}, a simulator primarily for autonomous cars, drones and robotics research. It is an extension of the original AirSim framework~\cite{shah2018airsim}, initially developed by Microsoft Research to provide a realistic simulation environment for training and testing autonomous systems. Based on Unreal Engine 5 (UE5)~\cite{unreal2025}, the simulator provides high-quality graphics and physics simulation. Additionally, Matlab and Python APIs are provided to allow simulation-based research without requiring knowledge of the simulator's internals. ASVSim also supports procedural generation (PCG) of environments (See Sec.~\ref{sec:pcg}). Because UE5 is used as a basis, we can leverage the blueprints system to easily change the simulation environments without writing code. With this visual scripting language, researchers can easily modify various aspects of the simulation, such as the objects in the waterway. Adjustments can include changing weather or time of day, altering the size, speed, orientation, and behavior of vessels, as well as changing camera perspectives to suit specific scenarios. Furthermore, it is straightforward to manipulate the environment to include new static and dynamic obstacles or change the layout. Researchers can also integrate custom 3D models or pre-existing assets from the Fab\footnote{https://www.fab.com/}, allowing for the creation of realistic and diverse environments tailored to research needs. In the first version of ASVSim, we have included three different basic example environments, a lake environment, a port environment, and a canal environment. These environments are shown in Fig.~\ref{fig:sim_envs}. Since ASVSim is built on UE5~\cite{unreal2025}, the hardware requirements follow those of the engine. The minimum requirements are a quad-core CPU, 16 GB RAM, and a GPU with 4 GB VRAM. The recommended specifications are a quad-core CPU (2.5 GHz or higher), 32 GB RAM, and a GPU with 8 GB VRAM. ASVSim supports Windows, macOS, and Linux. ASVSim also supports a headless mode, which disables rendering and allows the simulator to run without a display. This is particularly useful for training RL agents or collecting non-visual data, as it significantly reduces computational overhead.

ASVSim extends upon the work of~\cite{cosysairsim2023jansen} (see Fig.~\ref{fig:sim_overview}) by adding vessel dynamics (see Sec.~\ref{sec:dynsim}), including the modeling of disturbances such as currents and wind. We provide multiple vessel models, based on existing research vessels, together with inland waterway environments. Furthermore, we include radar simulation, as explained in Sec.~\ref{sec:radsim}. Finally, we extended the Python/Matlab API to include functionalities for controlling the vessel, changing disturbances, and retrieving sensor data (see Sec.~\ref{sec:vescontrol}).  We also provide documentation\footnote{https://bavolesy.github.io/idlab-asvsim-docs/} on the simulator and how to install, run and create your own projects. In Section~\ref{sec:dataset}, we highlight the annotated dataset generation functionality and in Section~\ref{sec:vescontrol} we demonstrate how to control the simulator, both using the API.

\subsection{Dataset Generation}\label{sec:dataset}

Vision-based perception is essential for autonomous shipping systems in constrained inland waterway and port scenarios. Although computer vision techniques have been extensively applied in areas such as manufacturing and autonomous driving~\cite{zhou2022computer,kiran2021deep}, progress in the maritime domain has been limited by the relative lack of annotated inland waterway vision datasets~\cite{su2023survey,trinh2024survey}. Recent research~\cite{bu2021carla} has shown that synthetic camera data generated from UE5 can significantly enhance the performance of object detection models on real-world data. To address this gap in the ocean surface robotics community, we enable dataset generation for object detection, depth estimation, and image segmentation with their corresponding ground-truth labels.

ASVSim currently includes vessel blueprints for a large container vessel and a fisher boat, along with photorealistic water rendering. Researchers can import their own 3D models (e.g., custom vessel types, port infrastructure) or use assets from the Fab marketplace. Additionally, plugins such as Cesium for Unreal\footnote{https://cesium.com/platform/cesium-for-unreal/} can be used to stream real-world photorealistic 3D environments (e.g., Google Photorealistic 3D Tiles), enabling the creation of geographically accurate waterway and port scenarios without manual environment modeling. Integration of Cesium for Unreal into ASVSim is currently a work in progress. Ground-truth annotations are generated automatically by ASVSim through UE5's built-in semantic segmentation: each object class in the scene is assigned a unique color in the segmentation map, from which pixel-wise labels, bounding boxes, and instance masks are derived programmatically via provided Python scripts. The semantic class definitions are configured by the user; in Section~\ref{sec:cv} we use three classes (sky, water, and obstacles), but researchers can define their own taxonomy to suit their application. This eliminates the need for manual annotation and ensures consistency across generated datasets.

\begin{figure}[t!]
    \centering
    \subfigure[]{\includegraphics[width=0.48\columnwidth]{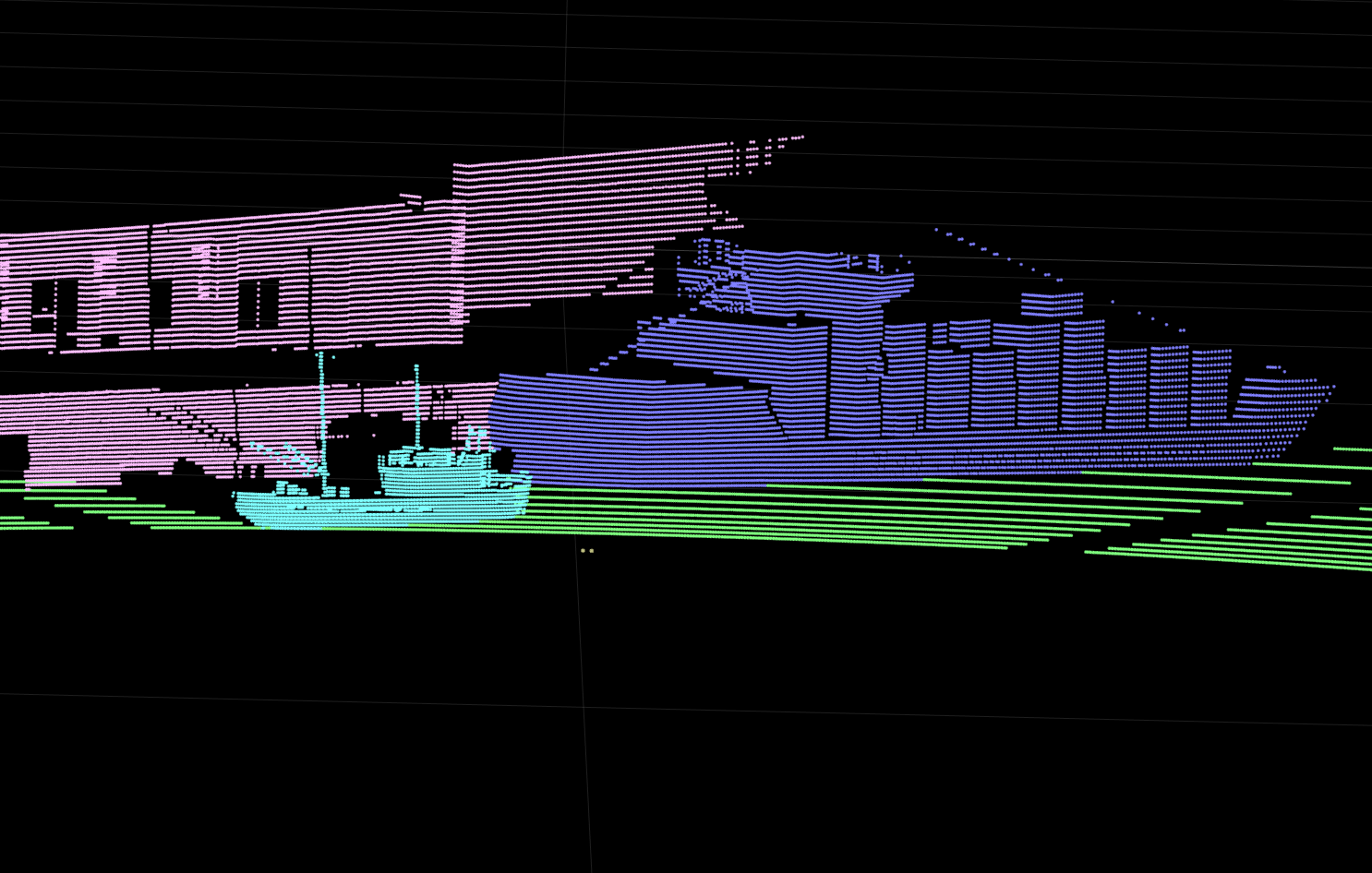}}
    \subfigure[]{\includegraphics[width=0.48\columnwidth]{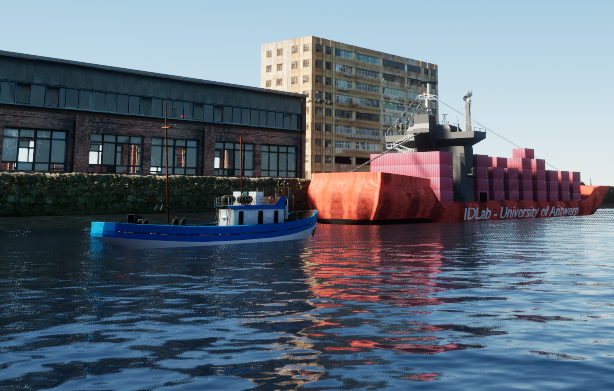}}
    \caption{\textbf{(a) Generated point cloud, (b) corresponding camera image.}}
    \label{fig:pointcloud}
\end{figure}

Moreover, as ASVSim provides access to various sensors, such as LiDAR, radar, and depth camera, these datasets can also be used for sensor fusion tasks, which are increasingly important for advancing autonomy. For example, the depth camera captures precise distance measurements for each pixel in the image, which can be used to develop monocular depth estimation models~\cite{Spencer_2023_CVPR}. These models are crucial for scenarios where only RGB cameras are available, enabling the estimation of distances to objects in the scene based solely on visual input~\cite{lee2021semantic}. Furthermore, combining depth estimation from RGB cameras with traditional radar systems allows for the detection of smaller objects that might be missed by radar alone due to low sparsity or quality~\cite{ophoff2019exploring}. This enhanced optical precision is particularly important in IWT, where the presence of small obstacles, such as buoys, poses significant challenges for safe and reliable autonomous navigation. Furthermore, the simulator generates ground-truth labels for LiDAR point clouds, supporting dataset creation for 3D LiDAR-based object detection research, relevant for precise maneuvering such as docking or tugging. Fig.~\ref{fig:cv_vessels} demonstrates the generated dataset, showing vessels detected by the semantic camera with depth camera measurements and automatically generated bounding boxes based on the semantic color mapping, along with the distance to these vessels. These images were collected using the \textit{simGetImages()} command from the Python API. Additionally we generated a point cloud (see Fig.~\ref{fig:pointcloud}) using the available LiDAR sensor and the Matlab API, via the command shown in Listing~\ref{lst:airsim_matlab_api}. This point cloud also automatically provides an instance segmentation ground truth. For more information on the LiDAR sensor, the reader is referred to~\cite{jansen2022physical}.

\subsection{Procedural Environment Generation}\label{sec:pcg}
\begin{figure}[t!]
    \centering
    \includegraphics[width=0.48\columnwidth]{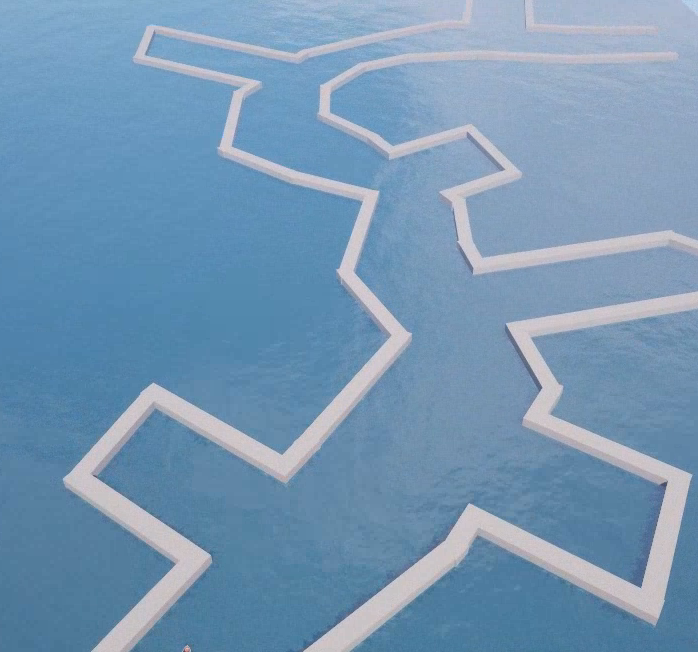}
    \includegraphics[width=0.48\columnwidth]{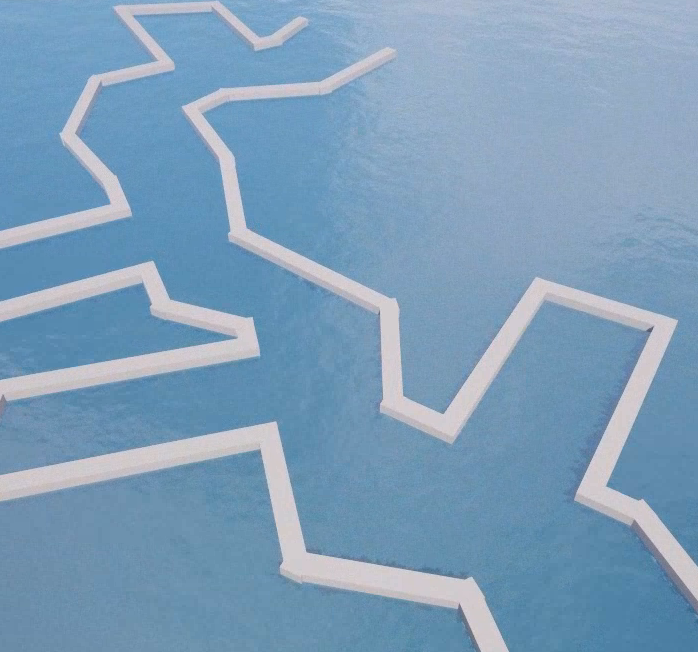}
    \caption{\textbf{Example port-like waterways generated in ASVSim using the Python API. This environment can be used for training of RL agents for autonomous vessel navigation.}}
    \label{fig:pcg}
\end{figure}

Recently, RL for autonomous navigation has gained significant interest. However, its success is often limited by the diversity of scenarios encountered during training. When trained in only a few environments, RL tends to struggle with generalizing to unseen situations. A common strategy to address this limitation is domain randomization~\cite{tobin2017domain}. By randomizing the environments and obstacles during training, RL may generalize better to unseen scenarios. Using the API, we can procedurally generate randomized port-like environments. When creating an environment, parameters such as the number of segments, random seed, width range, and the angle range between segments can be specified. In Fig.~\ref{fig:pcg}, some examples of port-like environments can be seen. These can be generated at runtime, allowing for flexible RL training.

\subsection{Controlling Vessels}\label{sec:vescontrol}

Besides receiving the point cloud, Listing~\ref{lst:airsim_matlab_api}, provides a minimal example of how to control a vessel using the Matlab API. First, a connection to the running simulator is established. After the connection has been set up, measured state information can be received from a simulated Global Navigation Satellite System (GNSS) sensor. This includes information such as the current location, and velocities on the surge and sway axes. It is also possible to receive the acceleration vector of the vessel, using the available Inertial Measurement Unit (IMU) (allowing for Bayesian state estimation methods). More information about the ego vessel and its environment, such as wind speed and direction, the current time of day and the weather conditions are all available via the API. All sensor data, such as camera, LiDAR, radar and sonar sensors can also be read out. 

\begin{lstlisting}[language=Matlab, caption=example on thrust control with a PID using the Matlab API., label={lst:airsim_matlab_api}, breaklines=true] 
    client = AirSimClient();
    % Get LiDAR pointcloud p and semantic labels l
    [p, l, ~, ~] = client.getLidarData('Lidar1', true, '');
    % Define PID with (Kp, Ki, Kd)
    pid = SimplePID(1.5, 1.0, 0.2);
    desired_speed = 0.51  % in m/s          
    % Control loop
    while true
        [gnssData, ~, ~] = client.getGpsData("Gps",'');
        v = gnssData.velocity;
        speed = sqrt(v(1).^2 + v(2).^2)
        thrust = pid.compute(desired_speed, speed)    
        client.setVesselControls(thrust, 0.5, '');
        pause(0.1);
    end
\end{lstlisting}

For an exhaustive summary of all API availability, we refer the reader to the documentation. With this sensor information, it is possible to manually control the vessel or use autonomous navigation algorithms to control all thrusters of the vessel. In the example, we set the thrust to 0.8 and the angle to 0.5 (both are normalized from 0 to 1, where an angle of 0.5 is equal to 0° and 0 and 1 represent the maximum angle of the thruster in both directions). This will make the vessel move forward in the surge direction with 80\% of its maximum thrust. Additionally, non-controllable vessels can also be spawned, which are useful for testing the performance of navigational algorithms when other vessels are operating in the proximity of the ego vessel. Listing~\ref{lst:airsim_matlab_api} furthermore displays a control loop that employs a proportional-integral-derivative (PID) controller to control the normalized thrust force on the (single-thruster) vessel. In this example case the PID is set to reach and hold a surge velocity of 1 knot.

\section{Experiments}\label{sec:experiments}
To further demonstrate the potential of ASVSim for autonomy research, we include three experiments. As we also include the source code for these experiments, they serve as a starting ground for future research. First, in Sec.~\ref{sec:cv}, we evaluate an existing waterway segmentation model on synthetic data. Secondly, in Sec.~\ref{sec:pathplanning}, we perform two path planning experiments. The first experiment demonstrates a navigation approach, based on the Dynamic Window Approach (DWA)~\cite{fox1997dynamic}, to navigate the MilliAmpere ferry~\cite{mAFerry} in a restricted waterway. The second experiment trains an RL agent to navigate the ferry through procedurally generated port channels with both static and dynamic obstacles.

\subsection{Waterway Segmentation}\label{sec:cv}

\Figure[t!](topskip=0pt, botskip=0pt, midskip=0pt)[width=1.0\textwidth]{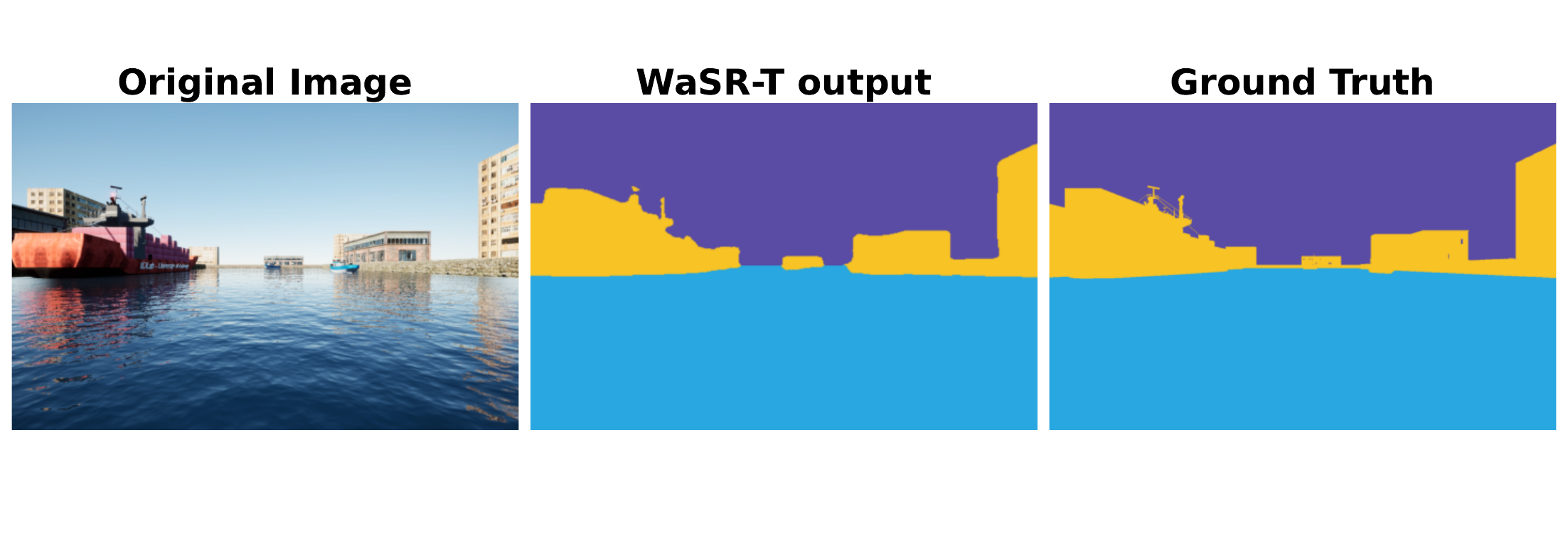}
{\textbf{A representative example output of the WaSR-T model~\cite{vzust2022temporal}. From left to right: the original simulated image, the WaSR-T output, and the ground-truth segmentation map. Blue denotes the waterway, purple denotes sky, yellow denotes all obstacles.}\label{fig:segmentation}}

In addition to the ability of collecting data (See Section~\ref{sec:dataset}), we investigate if an existing computer vision model (trained on real data) also works on the simulated camera images. For this, we perform a limited experiment that employs the WaSR-T~\cite{vzust2022temporal} segmentation model in ASVSim. Importantly, we use a model that was pre-trained exclusively on real-world data and perform no additional training or fine-tuning. We only evaluate the model's performance on synthetic ASVSim images. The WaSR-T model is specifically trained for waterway and obstacle segmentation in an inland context and hence closely aligns with the setting of this work. For the experiment, a trajectory is collected in a simulated inland waterway, using a camera mounted at the bow of the ego vessel (milliAmpere ferry). The recorded video (downsampled to 2 frames per second) is then segmented by the deep learning model. Interestingly, WaSR-T succeeds in the segmentation task, even on simulated ASVSim data, as demonstrated by the results in Table~\ref{table:segmentation} and further supported by the example in Fig.~\ref{fig:segmentation}. The model achieves a very high test score on segmenting the water class (all close to 0.99). Additionally, the model almost fully succeeds in differentiating between the sky and obstacles, however, the model sometimes does not capture all minor details, such as strips of air that are visible through windows of buildings. This good performance of WaSR-T on the synthetic data demonstrates the potential of ASVSim for preliminary validation of computer vision models during research, and suggests that the simulator could indeed be used for efficient pre-training of deep learning models. However, we leave a more elaborate investigation of these aspects for future work, as results are expected to be highly dependent on how visually realistic the UE5 environments are made.

\begin{table}
\centering
\caption{\textbf{Average test metrics of the WaSR-T model~\cite{vzust2022temporal} during a simulated test trajectory. Intersection over Union (IoU), Precision, and Recall for Water, Air, and Obstacle classes are shown. Precision and recall are calculated by considering the segmentation task as a pixel-wise classification problem.}}
\begin{tabular}{|c|c|c|c|}
\hline
    Metric & Water & Air & Other \\
\hline
    IoU & 0.9916 & 0.9859 & 0.9418 \\
    Precision & 0.9971 & 0.9973 & 0.9806 \\
    Recall & 0.9945 & 0.9885 & 0.9598 \\
\hline
\end{tabular}
\label{table:segmentation}
\end{table}

\subsection{Path Planning}\label{sec:pathplanning}

Within the field of autonomous navigation, path planning is the process of determining a feasible path for an object, in this case a vessel, to navigate from a starting point to a goal location. The general problem of path planning can be divided into two main categories: global and local path planning. Global path planning algorithms find a static path from the starting position to a goal, whilst local path planning algorithms aim to navigate safely between two waypoints on that global path. The determined path must avoid collisions with obstacles. For inland shipping, this becomes increasingly complex as the vessel must take into account various constraints, such as fuel consumption or regulatory and physical constraints. ASVSim has functionality for researchers to implement and test both traditional navigation methods and RL algorithms for path planning. We demonstrate this functionality in Sections~\ref{sec:dwa} and~\ref{sec:rl}.

\subsubsection{Autonomous Navigation with DWA}\label{sec:dwa}
As ASVSim is focused on IWT navigation, we consider a local path planning experiment, assuming that a global path is provided. It is important to distinguish between global and local path planning methods: commonly cited methods such as $A^*$~\cite{hart1968formal} and $RRT^*$~\cite{karaman2011sampling} are global planners that compute a complete, static path from start to goal, but do not account for vehicle dynamics during execution. For the local planning layer considered here, dynamically-feasible methods are more appropriate. Alternative local planners include Model Predictive Control (MPC)~\cite{zheng2014trajectory} and sampling-based approaches such as Model Predictive Path Integral (MPPI)~\cite{williams2016aggressive}, which can handle nonlinear dynamics and constraints. These methods typically require online optimization or Monte Carlo rollouts, incurring higher computational overhead which can hinder real-time operation. ASVSim can support any of these planners through its API. For this experiment, DWA~\cite{fox1997dynamic} was selected for its lower computational complexity, its established use in USV navigation and collision avoidance~\cite{SERIGSTAD20181, kim2021collision}, and because it directly encodes the vessel's dynamic constraints through the concept of a dynamic window and integrates naturally with the radar-based obstacle map and vessel state (heading, velocity) that ASVSim provides natively via its sensors. The approach works by creating a window of possible linear and angular velocities whilst taking into account the current dynamics of the vessel. All admissible velocity pairs are then evaluated with a predefined cost function over a trajectory with a finite horizon. To calculate admissible trajectories, the dynamics model of the vessel is used. The cost function that was used in the experiment is displayed in~\eqref{eq:costfunction}. To implement DWA, we opted for the PythonRobotics library~\cite{sakai2018pythonrobotics}. We integrated the algorithm with the simulator and tested it on a channel-navigation scenario with a number of obstacles. The trajectory begins at (0, 0) m and proceeds through 11 global waypoints to approximately (580, 240) m. A waypoint is considered reached when the vessel is within 10 m of the goal, after which the next goal is activated. A radar (see Sec.~\ref{sec:radsim}) was used to detect the edges of the channel and other obstacles (such as moored vessels). The radar scan is then used to create an obstacle map, which in turn is used to calculate the cost function for each trajectory in the dynamic window. The trajectory with the lowest cost is then chosen and executed. 
The cost function is defined as follows:
\begin{equation}\label{eq:costfunction}
    C_{total} = C_{speed} + C_{obstacle} + C_{goal},
\end{equation}
where \(C_{speed}\) is the cost related to the speed with which the vessel travels, \(C_{obstacle}\) is the cost of how close the vessel is to obstacles, and \(C_{goal}\) is the cost of how close the vessel's heading is to the desired goal heading. These costs are calculated as follows:
\begin{equation}
    \begin{aligned}
        C_{goal} &= G_{goal} \cdot \arctan\left(\frac{\sin(\psi_{goal} - \psi_{vessel})}{\cos(\psi_{goal} - \psi_{vessel})}\right) \\  
        C_{obstacle} &= G_{obstacle} \cdot \begin{cases} 
            \frac{1}{d_{\min}} & \text{if } d_{\min} > d_{threshold} \\
            +\infty & \text{otherwise}
        \end{cases} \\
        C_{speed} &= G_{speed} \cdot (v_{\max} - v_{vessel}), \\
    \end{aligned}
\end{equation}

with \(G_{goal}\), \(G_{obstacle}\), and \(G_{speed}\) being constants to weigh the importance of each cost. \(d_{\min}\) represents the minimum distance to obstacles along the trajectory, while \(d_{threshold}\) denotes the vessel's dimensions. A collision is assumed when \(d_{\min} < d_{threshold}\), and the cost is set to \(+\infty\). \(\psi_{goal}\) is the desired heading to the goal, \(\psi_{vessel}\) is the heading of the vessel, and \(v_{\max}\) is the maximum velocity of the vessel. After manual tuning of the cost function and configuring the parameters, the vessel navigated the path without collisions. This scenario is visualized, along with the corresponding radar detection, in Fig.~\ref{fig:radar_example}. Although DWA navigates the channel successfully, it must be stated that the high computational cost is a significant downside of the methodology. This downside can be important for long-horizon problems such as vessel navigation, especially in busy IWT scenarios, where the situation can change significantly over a short period of time. 

\Figure[t!](topskip=0pt, botskip=0pt, midskip=0pt)[width=1.0\textwidth]{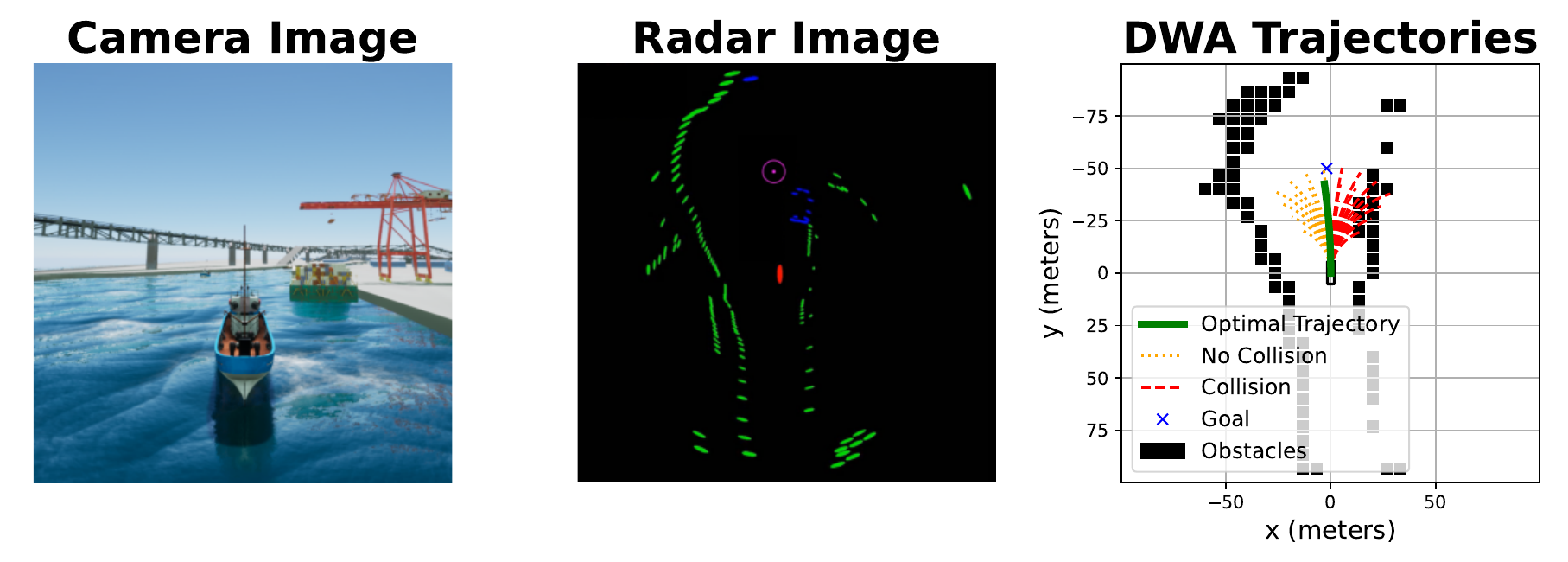}
{\textbf{Overview of the path planning experiment with DWA to navigate a vessel through a channel. From left to right: Camera image, radar image (with goal as purple circle), and simulated DWA trajectories.}\label{fig:radar_example}}

\begin{figure}[t!]
    \centering
    \includegraphics[width=\columnwidth]{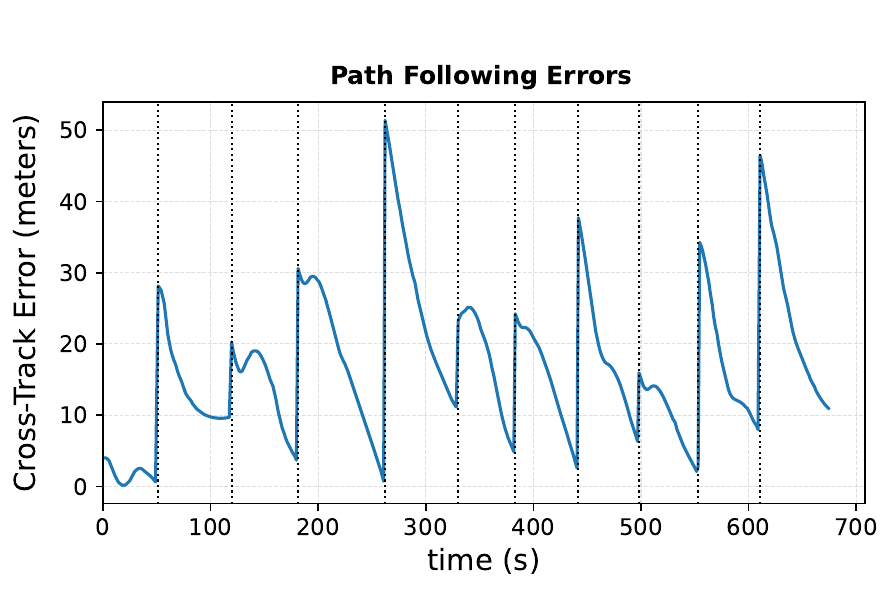}\\[-7pt]
    \includegraphics[width=\columnwidth]{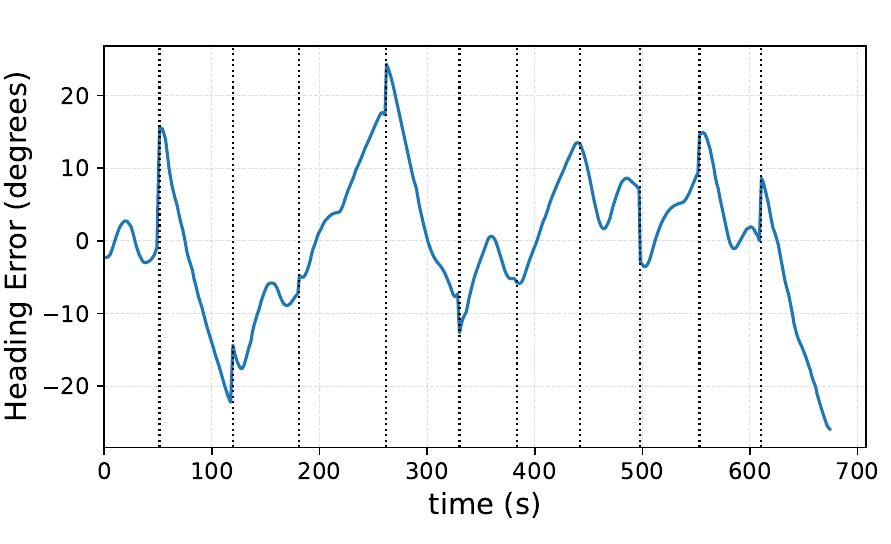}
    \caption{\textbf{Path following errors~\cite{lekkas2014minimization} for the DWA experiment. The dotted lines represent the timesteps in which a new point on the global path is set. Top: Cross-track error, Bottom: Heading error.}}
    \label{fig:dwa_error}
\end{figure}

To demonstrate that DWA successfully navigates the channel, Fig.~\ref{fig:dwa_error} reports the cross-track and heading errors~\cite{lekkas2014minimization} over time. Sharp increases in cross-track error occur immediately after a waypoint switch (indicated by the dotted lines), which is expected as we do not enforce arrival at the exact center of a waypoint with a particular orientation. These pose errors lead to temporary deviations, but DWA consistently demonstrates its ability to correct these errors as it navigates toward the next waypoint, shown by the decreasing cross-track error after each switch. We notice similar trends in the heading error, but with some oversteering. When DWA attempts to correct the heading, the vessel's angular momentum causes it to frequently overshoot the target orientation. This overshooting is visible in the graph where after a waypoint switch, the absolute value of the heading error decreases, crosses zero and then increases again. Increasing the planning horizon and the resolution of the dynamic window would likely reduce heading overshoot, and using a more accurate dynamics model for DWA instead of the linearized model we used could further mitigate oversteering. However, both approaches come with an increase in computation time. Note also that the ideal path does not account for obstacles and the vessel's maneuvers occasionally require unexpected turns, which further increase the cross-track and heading errors. The along-track error~\cite{lekkas2014minimization} is not shown on the figure, as it is not relevant for this specific scenario. 

\subsubsection{Reinforcement Learning}\label{sec:rl}
As mentioned in Section~\ref{sec:dwa}, a key limitation of DWA is its computational complexity, which scales poorly with the resolution of possible trajectories and trajectory length, with a median computation time of approximately 1.3\,s per planning step in our experiments (see Appendix~\ref{appendix:dwa_timing}). Another significant disadvantage of traditional control methods such as DWA is the need to manually adjust various parameters, in addition to the cost function, such as window size, time horizon, and velocity resolutions. To address these limitations, recent research has shown promising results using RL for path planning in autonomous shipping~\cite{zhang2024path, lesyRobustOfflineReinforcement2025a}. In this section, we demonstrate how ASVSim can be used to train RL agents for autonomous navigation in shipping applications.

RL~\cite{sutton1998reinforcement} is used to solve sequential decision-making problems where an agent learns an optimal strategy through interaction with an environment. These problems are modeled as Markov Decision Processes (MDPs). An MDP can be defined as a tuple $(S, A, P, R, \gamma)$, where \(S\) is the set of states, \(A\) is the set of actions, \(P\) is the transition probability function, which defines the probability of transitioning from one state to another given an action, \(R\) is the reward function, and \(\gamma\) is the discount factor, which balances immediate rewards with future returns. The agent's goal is to learn an optimal policy that maps states to actions to maximize the cumulative reward across the horizon $H$. The definition of the optimal policy can be seen in~\eqref{eq:optimalpolicy}.

\begin{equation}\label{eq:optimalpolicy}
    \pi^* =  \argmax_{\pi} \left[ \mathbb{E}_{a \sim \pi} \left( \sum_{t=0}^{H} \gamma^t R(s_t, a_t) \right)\right]
\end{equation}

Through continuous interaction with the environment, the agent learns an optimal policy by exploring different actions and updating its strategy based on received rewards. This continuous interaction can be problematic for existing systems, since learning policies on a real vessel can be time-consuming, costly and dangerous. Furthermore, for autonomous shipping solutions to achieve commercial viability, they must demonstrate robust generalization across different vessels and maritime environments. Our previous research~\cite{lesyRobustOfflineReinforcement2025a} has demonstrated that RL-based approaches maintain strong performance across vessels with varying physical properties, such as mass, without requiring retraining or extensive manual recalibration. This represents a significant advantage over traditional control methods, which typically require substantial domain expertise and manual tuning for each new vessel configuration. The success of RL algorithms heavily depends on training with diverse, representative data. ASVSim addresses this need by enabling the creation of varied training scenarios with dynamic obstacle configurations, different weather and water conditions, and various port and waterway layouts. This rich simulation environment allows for training of agents across a wide range of IWT scenarios.

To leverage the PCG system described in Section~\ref{sec:pcg}, we train an RL agent to navigate the procedurally generated port channels. The task is to steer the vessel from the starting position through the channel to a goal location. The terrain is regenerated every 10 episodes using a new random seed, producing a different port layout with varying channel geometry and border shapes each time. The \texttt{getGoal} API function is used to retrieve the goal coordinates for a given section of the generated port. By querying \texttt{getGoal} for each section from 1 up to the total number of generated sections, we obtain both intermediate waypoints and the final goal, which the agent must navigate to sequentially. The number of waypoints thus depends on the length of the generated environment. Additionally, at the start of every episode, static obstacles (buoys) and dynamic obstacles (vessels) are randomly spawned along the path between the vessel and the goal. Static buoys remain fixed throughout the episode, while dynamic vessels are assigned a random linear velocity and heading. This combination of terrain randomization, static buoys, and moving vessel traffic exposes the agent to a wide distribution of navigation scenarios during training, encouraging the learning of generalizable policies rather than the memorization of a single fixed environment. We included this experiment in the code release, along with the packaged environment, a trained model checkpoint, and an evaluation script, which can be used as a starting point for research towards RL agents in ASVSim.

\begin{figure}[t!]
    \centering
    \subfigure[]{\includegraphics[width=0.27\columnwidth]{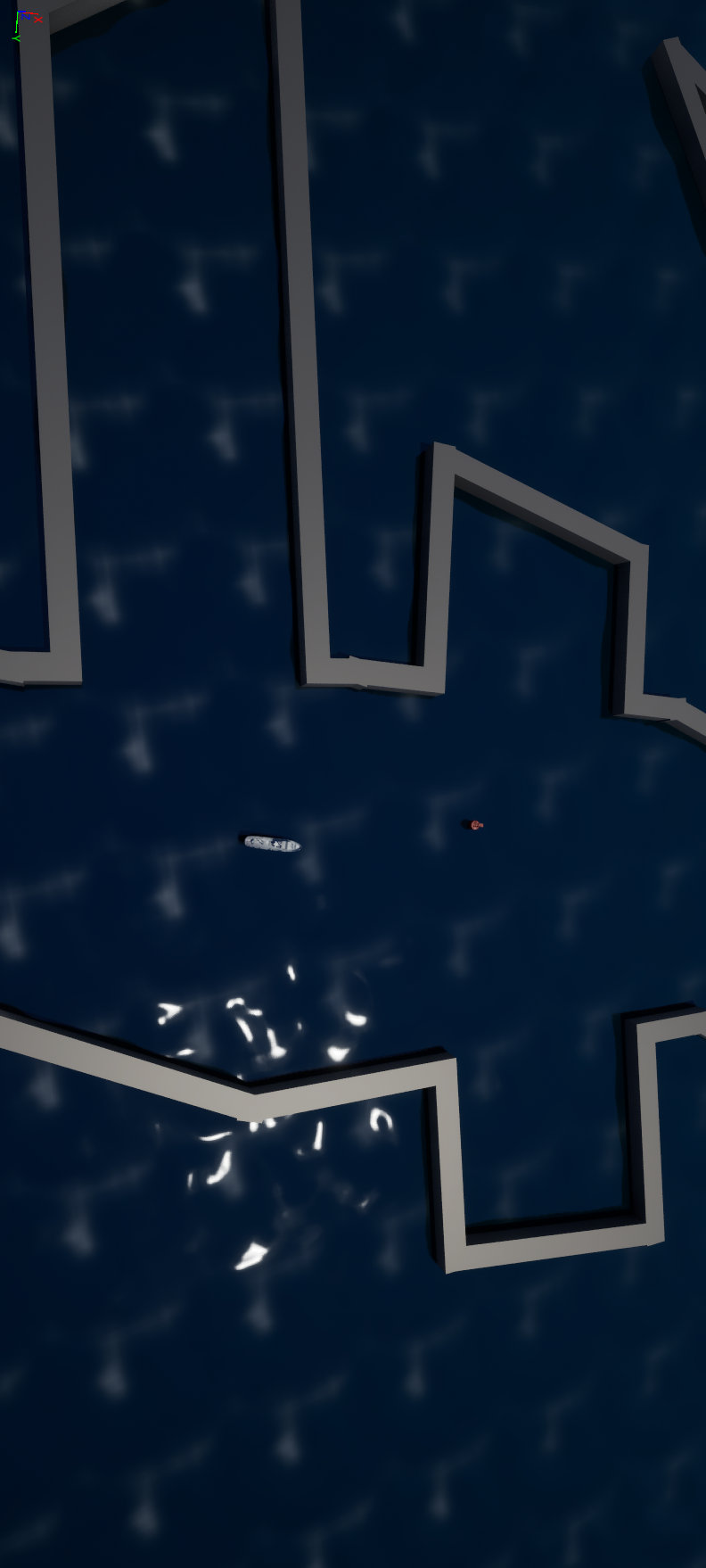}}
    \subfigure[]{\includegraphics[width=0.6\columnwidth]{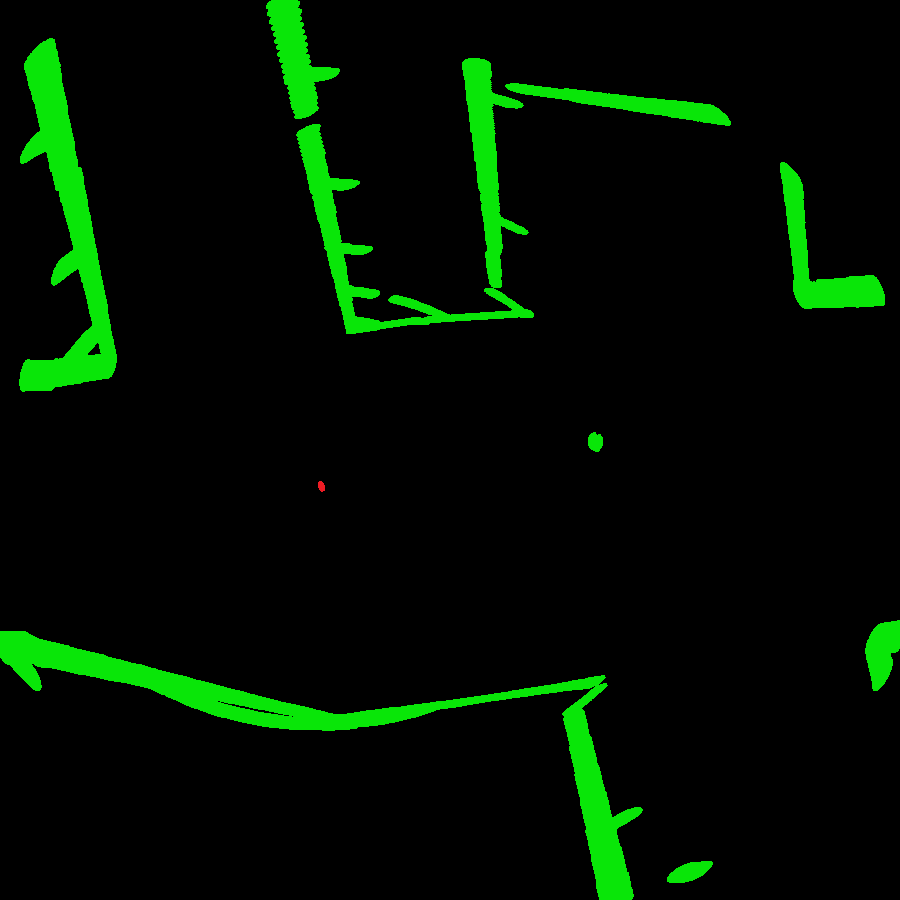}}
    \caption{\textbf{(a) Top-down simulator view of a procedurally generated port channel: the vessel navigates toward the goal through a channel with spawned obstacles (buoys and vessels). (b) The corresponding LiDAR point cloud as perceived by the agent, with detections shown in green and the vessel position in red.}}
    \label{fig:rl_env}
\end{figure}

In our experiment, at every timestep, the agent receives a 54-dimensional observation vector and outputs a continuous action, which is executed in the environment. The agent then receives the next observation and a scalar reward. This process repeats until the agent reaches the final goal, collides with an obstacle, or a maximum of 800 timesteps is exceeded. When an episode ends, the vessel is reset to its starting position and new obstacles are spawned. A waypoint is considered reached when the vessel is within 10~m of it, after which the next waypoint becomes the active navigation target.

The observation space consists of waypoint information $\boldsymbol{d}_{wp}$ (2D displacement vectors and scalar distances to the previous, current, and next waypoint), heading information $\boldsymbol{h}_\psi$ (heading error to the current waypoint, and vessel heading encoded as $\sin\psi$, $\cos\psi$), the linear velocity $\boldsymbol{\nu}$, linear and angular acceleration $\boldsymbol{\dot{\nu}}$, previous actions $a_{t-1}$, and LiDAR observations $\boldsymbol{d}_{lidar} \in \mathbb{R}^{36}$. The LiDAR sensor produces 3600 range measurements which are min-pooled into 36 sectors of 10° each. The full observation vector is:

\begin{equation}\label{eq:observation}
    \boldsymbol{\text{obs}}_t = \left[ \boldsymbol{d}_{wp},\, \boldsymbol{h}_\psi,\, \boldsymbol{\nu},\, \boldsymbol{\dot{\nu}},\, a_{t-1},\, \boldsymbol{d}_{lidar} \right]
\end{equation}

The vessel used in this experiment is the milliAmpere ferry, whose dynamics are described in Section~\ref{sec:dynsim}. The agent controls only the rear thruster, with a continuous 2-dimensional action space for thrust force (normalized to $[0, 0.7]$) and thruster angle (normalized to $[0.48, 0.52]$), with the same controls as in Section~\ref{sec:vescontrol}. Both ranges are constrained to limit ineffective and unstable actions.

The reward function is designed to encourage steady progress toward the current waypoint while penalizing collisions and idle behavior. At each timestep, the agent receives a progress reward based on the change in Euclidean distance to the active waypoint, combined with a constant time penalty:

\begin{equation}\label{eq:rewardfunction}
    R(s_t, a_t) = \underbrace{\left( d_{t-1} - d_{t} \right)}_{\text{progress}} - \underbrace{0.1}_{\text{time penalty}}
\end{equation}

where $d_{t}$ is the Euclidean distance to the current active waypoint at timestep $t$. When an intermediate waypoint is reached, the navigation target advances to the next waypoint and the distance tracking resets accordingly. Terminal rewards are assigned at the end of an episode:

\begin{equation}\label{eq:terminalreward}
    R_{terminal} = \begin{cases}
        +500 & \text{if final goal reached} \\
        -100 & \text{if collision}
    \end{cases}
\end{equation}

The algorithm used for training is CrossQ~\cite{bhatt2024crossq}, a sample-efficient off-policy actor-critic method, as implemented in the SB3-Contrib extension of Stable Baselines3~\cite{stable-baselines3}. Observations are normalized using running statistics to stabilize training. The full set of hyperparameters is listed in Table~\ref{tab:rl_params} in Appendix~\ref{sec:rl_parameters}.

\Figure[t!](topskip=0pt, botskip=0pt, midskip=0pt)[width=0.99\columnwidth]{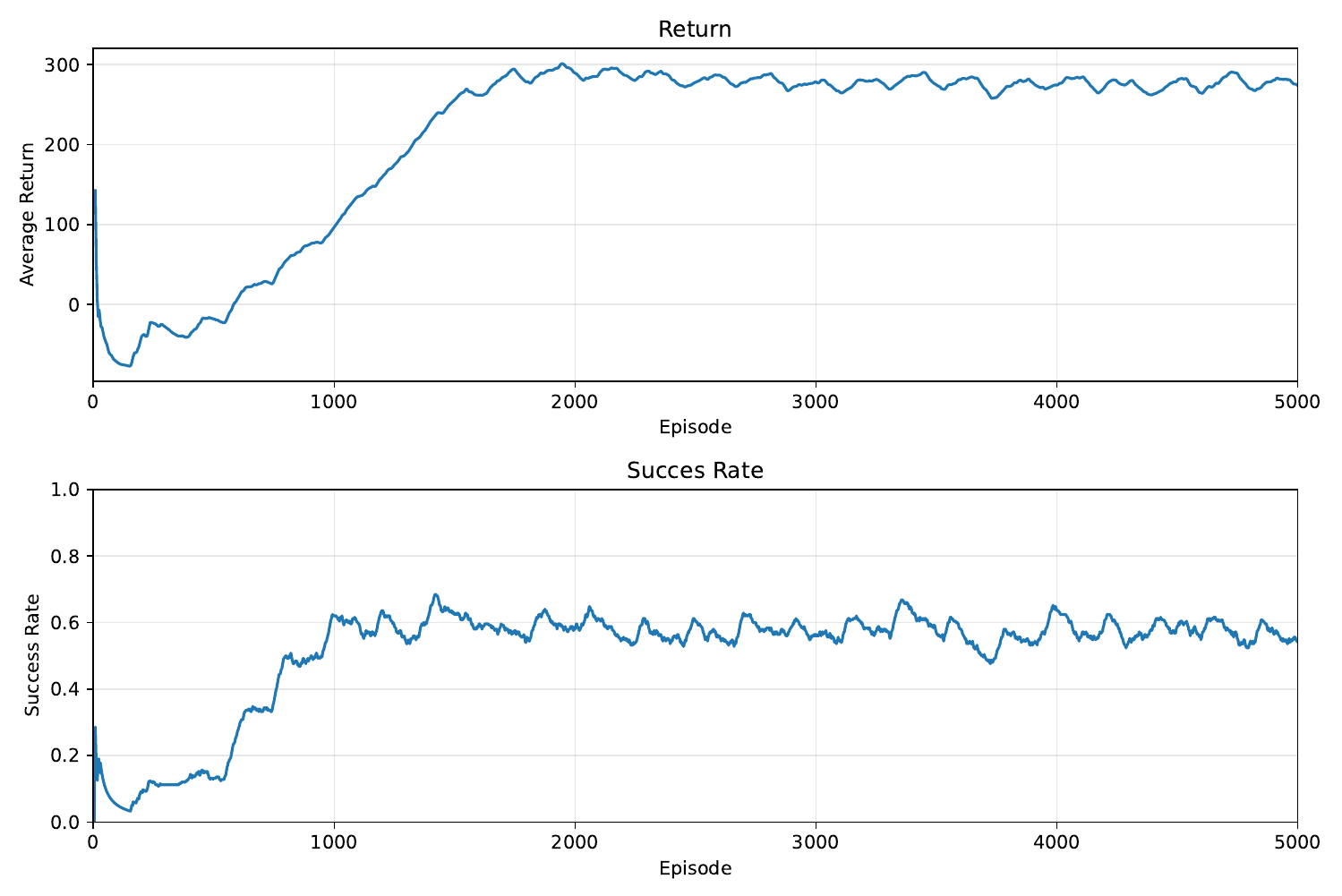}
{\textbf{Training curves for the CrossQ agent over 5000 episodes in procedurally generated port environments. Top: average return. Bottom: success rate (rolling average).}\label{fig:rl_training}}

Fig.~\ref{fig:rl_training} shows the training curves over 5000 episodes from a single training run. We define the \textit{success rate} as the fraction of episodes in which the agent reaches the final goal without collision, and the \textit{return} as the cumulative reward collected over an episode. The average return increases steadily during the first 2000 episodes and then stabilizes. The success rate rises from near zero to approximately 60\%, where it plateaus. The remaining failures are primarily due to collisions with randomly spawned obstacles that appear in close proximity to or directly in front of the vessel. Since the agent only controls the rear thruster and can only move forward, it has limited maneuverability to avoid obstacles in these configurations. We note that expanding the action space to include additional thrusters (e.g., bow thruster) or allowing reverse thrust would increase the agent's ability to avoid such collisions. Further optimization of the reward function and training hyperparameters may also improve performance, which we leave to future work. An example of the PCG environment used during training is shown in Fig.~\ref{fig:rl_env}.

To evaluate the generalization of the trained policy, we tested the final checkpoint on 25 previously unseen PCG environments (10 episodes each, 250 episodes total) with random obstacle configurations every episode not encountered during training. The agent achieved a success rate of 57.2\% and a collision rate of 42.8\%. This is consistent with the training success rate of approximately 60\%, indicating that the policy generalizes to unseen environments without significant performance degradation. The failure mode remains the same: collisions with obstacles that are spawned in close proximity to or directly ahead of the vessel, where the limited action space (rear thruster only, forward motion) leaves insufficient room to maneuver.

\section{Discussion and Future Work}\label{sec:future_work}
Our results demonstrate that the proposed open-source simulator provides a useful basis for further research towards autonomous vessel navigation. Sec.~\ref{sec:cv} has demonstrated that, in the tested simulation environment, a computer vision model that works on real-world data, can also work in ASVSim. This implies the direct potential many research directions such as investigating the impact of weather conditions on the computer vision model or comparing multiple models in a wide variety of situations. However, the question if we could train a model purely in the simulator and then deploy it in reality remains unanswered and is left for future work. Investigating this would require creating a large variety of simulation environments. In addition to the computer vision experiment, Sec.~\ref{sec:dwa} and Sec.~\ref{sec:rl} demonstrate how the simulator can be used to train and test path planning algorithms. This allows for a decrease in time to deployment of novel algorithms, compared to only performing real-world experimentation. A limiting factor remains that several assumptions and simplifications have been made (such as the use of a 3DoF model and the radar limitations discussed in Section~\ref{sec:radsim}), leaving a necessity for real-world validation before deployment. Accordingly, ASVSim is currently best suited for prototyping and training algorithms in inland and port scenarios, with real-vessel transfer remaining an active direction for future work. Lastly, the Python and Matlab API, combined with the provided example experiments ensure that research within ASVSim is readily accessible to the research community.

There are several directions for future development that would further enhance the simulator's capabilities. Our research group has recently conducted experimental trials with a USV equipped with various sensors such as a depth camera, LiDAR, and IMU. A key direction for future work involves performing system identification, similar to the work presented in~\cite{pedersen2019optimization}, on this USV to precisely determine the parameters corresponding to its dynamics model. This will enable us to create a digital model of our USV in ASVSim and to simulate its dynamics. Such precise modeling may allow for the development of navigation algorithms and training of RL agents in simulation that can be effectively transferred to the physical vessel with minimal performance degradation. Furthermore, we plan to create more detailed and expansive inland waterway environments to generate realistic datasets and to train RL agents to autonomously navigate in these environments. We also aim to add more vessel models for IWT applications, including 6-degree-of-freedom (6DoF) models. Lastly, another possible direction for future work is to develop more sensor models, including multi-beam echosounders and radar-based positioning systems, such as radar beacons (RACONs). By releasing our work in an open-source manner, we invite the broader research community to collaborate, contribute, and extend its capabilities. Through collaboration, we hope to accelerate innovation in inland autonomous navigation research.

\section{Conclusion}\label{sec:conclusion}
In this paper, we presented ASVSim, an open-source simulation framework specifically designed for autonomous shipping research. This simulator addresses a current gap in the research domain, where no open-source simulators with high visual fidelity are available. By combining vessel dynamics models with relevant sensor simulations, we provide an extensive platform for developing and evaluating autonomous technologies for inland shipping applications. The experiments conducted on waterway segmentation and path planning demonstrate ASVSim's effectiveness in generating realistic sensor data and modeling vessel dynamics in inland waterway environments. 

\section{Acknowledgment}
During the preparation of this work the authors used Claude Sonnet 4 in order to help with latex commands and proofread the work. After using this tool, the authors reviewed and edited the content as needed and take full responsibility for the content of the published article.

\appendices

\section{\break DWA Implementation Details}\label{sec:dwa_parameters}
This section provides the implementation details for the DWA experiment described in Section~\ref{sec:dwa}. The implementation is based on the PythonRobotics library~\cite{sakai2018pythonrobotics}, which uses a linear motion model to forward-simulate candidate trajectories. This approximation is valid for sufficiently small time steps ($\Delta t = 0.1$\,s in our configuration), where the linearization error remains negligible.

Table~\ref{tab:dwa_params} lists all DWA parameters used in the experiments.

\begin{table}[h]
\centering
\caption{DWA planner configuration parameters. Note: these values were manually tuned for the DWA search space and do not represent the physical limits of the MilliAmpere ferry.}
\label{tab:dwa_params}
\begin{tabular}{llrl}
\hline
\textbf{Parameter} & \textbf{Symbol} & \textbf{Value} & \textbf{Unit} \\
\hline
\multicolumn{4}{l}{\textit{Velocity limits}} \\
Maximum speed          & $v_\text{max}$              & 25.0  & m/s     \\
Minimum speed          & $v_\text{min}$              & 1.0   & m/s     \\
Maximum yaw rate       & $\dot{\psi}_\text{max}$     & 30    & deg/s   \\
\hline
\multicolumn{4}{l}{\textit{Acceleration limits}} \\
Maximum acceleration   & $a_\text{max}$              & 20.0  & m/s$^2$ \\
Maximum yaw acceleration & $\ddot{\psi}_\text{max}$  & 30    & deg/s$^2$ \\
\hline
\multicolumn{4}{l}{\textit{Sampling resolution}} \\
Velocity resolution    & $\Delta v$                  & 0.3   & m/s     \\
Yaw rate resolution    & $\Delta \dot{\psi}$         & 1.0   & deg/s   \\
\hline
\multicolumn{4}{l}{\textit{Prediction}} \\
Time step              & $\Delta t$                  & 0.1   & s       \\
Prediction horizon     & $T_p$                       & 4.0   & s       \\
\hline
\multicolumn{4}{l}{\textit{Cost weights}} \\
Goal direction weight  & $G_\text{goal}$             & 0.5   & --      \\
Speed weight           & $G_\text{speed}$            & 0.2   & --      \\
Obstacle weight        & $G_\text{obstacle}$         & 1.0   & --      \\
\hline
\multicolumn{4}{l}{\textit{Vessel geometry (bounding box)}} \\
Vessel width           & $W$                         & 4   & m       \\
Vessel length          & $L$                         & 8   & m       \\
Goal check radius      & $r_\text{goal}$             & 10   & m       \\
\hline
\end{tabular}
\end{table}

\subsection{Search Space}

The dynamic window is bounded by both the kinematic limits of the vessel and the reachable velocities from the current state. In the worst case (full window), the number of candidate trajectories is given by:
\begin{equation}
    N_\text{candidates} = \left\lfloor \frac{v_\text{max} - v_\text{min}}{\Delta v} \right\rfloor \times \left\lfloor \frac{2 \, \dot{\psi}_\text{max}}{\Delta \dot{\psi}} \right\rfloor = 80 \times 60 = 4{,}800
\end{equation}
Each trajectory consists of $T_p / \Delta t = 40$ predicted states, yielding up to $192{,}000$ state evaluations per planning step. In practice, the acceleration constraints reduce the admissible window, particularly near the velocity and yaw rate boundaries.

\subsection{Computation Time}
\label{appendix:dwa_timing}

In our experiments (single-threaded Python with NumPy on an Intel Core i7-1260P, 32\,GB DDR5, Windows~10), the median per-step planning time was approximately 1.3\,s, with a typical range of 0.65--1.5\,s. Peak times exceeding 2\,s occurred when the vessel operated near the velocity or yaw rate boundaries, where the full admissible window is explored. A control frequency of 1\,Hz is commonly used for autonomous vessel navigation, making the current single-threaded implementation not feasible for real-time operation. However, since the trajectory evaluations are independent, GPU acceleration or vectorized batch computation could reduce planning times sufficiently for real-time deployment. Alternatively, coarsening the sampling resolution or shortening the prediction horizon would directly reduce the number of candidate evaluations at the cost of solution quality.

\section{\break RL Implementation Details}\label{sec:rl_parameters}
This section provides the hyperparameters for the RL experiment described in Section~\ref{sec:rl}. Table~\ref{tab:rl_params} lists all training, environment, sensor, and reward parameters. The simulation runs in real-time with no speed-up or parallelization; the total wall-clock time for training was approximately 42 hours.

\begin{table}[h]
\centering
\caption{CrossQ training and environment configuration parameters.}
\label{tab:rl_params}
\begin{tabular}{llrl}
\hline
\textbf{Parameter} & \textbf{Symbol} & \textbf{Value} & \textbf{Unit} \\
\hline
\multicolumn{4}{l}{\textit{Training}} \\
Learning rate          & $\alpha$                    & $3 \times 10^{-4}$ & --  \\
Discount factor        & $\gamma$                    & 0.99               & --  \\
Batch size             & $B$                         & 256                & --  \\
Replay buffer size     & $|\mathcal{D}|$             & $5 \times 10^5$    & --  \\
Learning starts        & --                          & 5000               & steps \\
Train frequency        & --                          & 1                  & steps \\
Total timesteps        & --                          & $2.5 \times 10^6$  & steps \\
Network architecture   & --                          & MLP, $2 \times 512$, ReLU & -- \\
\hline
\multicolumn{4}{l}{\textit{Environment}} \\
Terrain regen interval & --                          & 10                 & episodes \\
Static obstacles       & --                          & 4                  & --  \\
Dynamic obstacles      & --                          & 2                  & --  \\
Control frequency      & --                          & 4               & Hz  \\
Max timesteps          & $T_\text{max}$              & 800                & steps \\
Waypoint reach radius  & $r_\text{wp}$               & 10                 & m   \\
\hline
\multicolumn{4}{l}{\textit{LiDAR sensor}} \\
Channels               & --                          & 8                  & --  \\
Measurements per cycle & --                          & 450                & --  \\
Range                  & --                          & 1000               & m   \\
Horizontal FOV         & --                          & $[-180,\, 180]$    & deg \\
Vertical FOV           & --                          & $[-10,\, -2]$      & deg \\
\hline
\multicolumn{4}{l}{\textit{Action space}} \\
Thrust range           & --                          & $[0,\, 0.7]$       & --  \\
Thruster angle range   & --                          & $[0.48,\, 0.52]$   & --  \\
\hline
\multicolumn{4}{l}{\textit{Reward}} \\
Progress weight        & --                          & 1.0                & --  \\
Time penalty           & --                          & 0.1                & --  \\
Goal reward            & $R_\text{goal}$             & $+500$             & --  \\
Collision penalty      & $R_\text{collision}$        & $-100$             & --  \\
\hline
\end{tabular}
\end{table}

\subsection{Runtime Performance}\label{sec:fps}

Table~\ref{tab:fps} reports the performance of ASVSim during the RL experiment (Section~\ref{sec:rl}), with the RL training process running concurrently on the same machine. Measurements were taken with the LiDAR sensor active (450 measurements per cycle, 8 channels, $3{,}600$ points per second) at default graphics quality.

\begin{table}[h]
\centering
\caption{ASVSim performance during RL training on different hardware configurations and default graphics.}
\label{tab:fps}
\begin{tabular}{lllr}
\hline
\textbf{GPU} & \textbf{RAM} & \textbf{CPU} & \textbf{FPS} \\
\hline
NVIDIA RTX 2060 Super  & 64\,GB DDR4 & Intel i5-9600KF & 60  \\
NVIDIA T550            & 32\,GB DDR5 & Intel i7-1260P & 40  \\
NVIDIA RTX 3060 (Laptop)        & 16\,GB DDR5 & Intel i7-12700H & 80  \\
\hline
\end{tabular}
\end{table}

\bibliographystyle{IEEEtran}
\bibliography{references}

\begin{IEEEbiography}[{\includegraphics[width=1in,height=1.25in,clip,keepaspectratio]{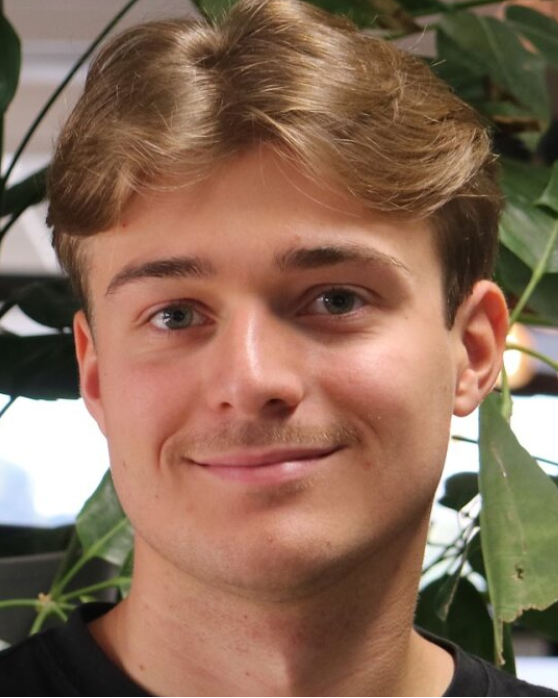}}]{Bavo Lesy}
received the bachelor’s and master’s degrees in applied electronics and ICT engineering from the University of Antwerp, in 2022 and 2023, respectively. Since 2023, he has been a Ph.D. Researcher within the IDLab research group at imec and the University of Antwerp. His research interests include (safe) reinforcement learning, autonomous vehicles and representation learning. 
\end{IEEEbiography}

\begin{IEEEbiography}[{\includegraphics[width=1in,height=1.25in,clip,keepaspectratio]{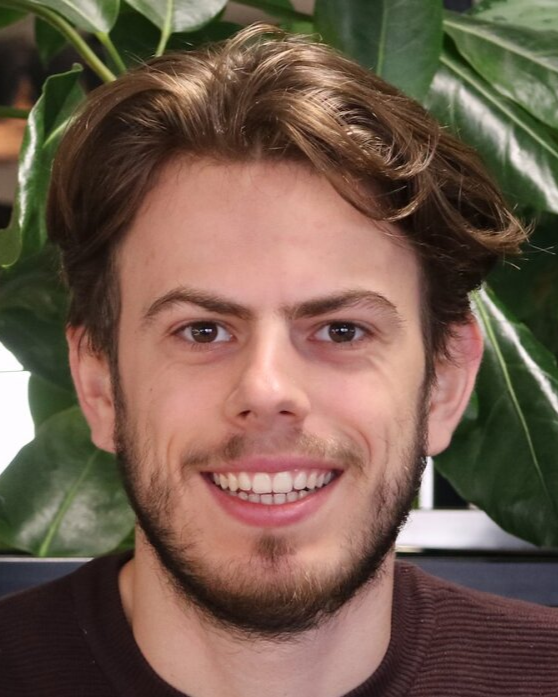}}]{Siemen Herremans} 
received the bachelor’s and master’s degrees in applied electronics and ICT engineering from the University of Antwerp in 2020 and 2021, respectively. Since 2022, he has been a Ph.D. Researcher within the IDLab research group at imec and the University of Antwerp. His research interests include robust (model-based) reinforcement learning, autonomous navigation and deep learning. 
\end{IEEEbiography}

\begin{IEEEbiography}[{\includegraphics[width=1in,height=1.25in,clip,keepaspectratio]{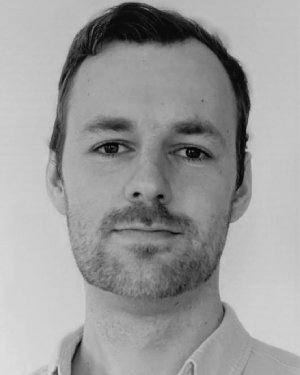}}]{Robin Kerstens}
received the B.Sc. degree in applied engineering electronics and ICT from the University of Antwerp, Antwerp, Belgium, in 2014, with a thesis focus on radar signal processing, the M.Sc.Eng. degree in electronics and ICT engineering technology from the University of Antwerp, with a thesis focus on sonar beamforming, and the Ph.D. degree in advanced array signal processing for in-air sonar. After a short spell in the automotive industry designing pressure sensors, he completed his Ph.D. degree. His research interests include in-air sonar, signal processing, radar, wave propagation, acoustics, machine learning, and algorithm development.
\end{IEEEbiography}

\begin{IEEEbiography}[{\includegraphics[width=1in,height=1.25in,clip,keepaspectratio]{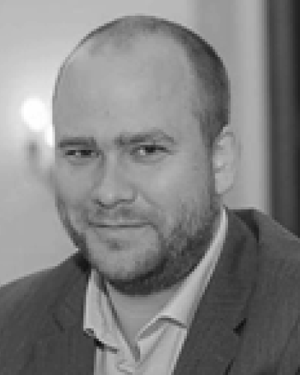}}]{Jan Steckel}
received the degree in electronic engineering from the University College “Karel de Grote,” Hoboken, Belgium, in 2007, and the Ph.D. degree from the Active Perception Laboratory, University of Antwerp, Antwerp, Belgium, in 2012, with a dissertation titled “Array processing for in-air sonar systems-drawing inspirations from biology”. During this period, he developed state-of-the-art sonar sensors, both biomimetic and sensor-array based. During his postdoctoral period, he was an active member of the Center for Care Technology, University of Antwerp, where he was in charge of various healthcare-related projects concerning novel sensor technologies. Furthermore, he pursued industrial exploitation of the patented 3-D array sonar sensor, which was developed in collaboration during the Ph.D. degree. In 2015, he became a Tenure Track Professor with the Cosys-Laboratory, University of Antwerp, where he researches sensors, sensor arrays, and signal processing algorithms using an embedded, constrained systems approach.
\end{IEEEbiography}

\begin{IEEEbiography}[{\includegraphics[width=1in,height=1.25in,clip,keepaspectratio]{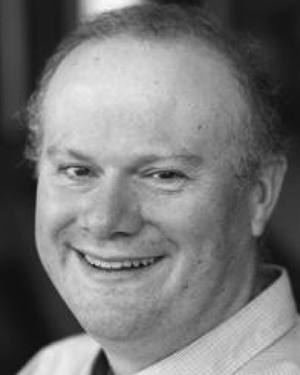}}]{Walter Daems}
(Senior Member, IEEE) received the B.Sc. degree in electronics from the Katholieke Industriële Hogeschool Antwerp, Hoboken, Belgium, in 1994, and the M.Sc. and Ph.D. degrees in electrical engineering from the Katholieke Universiteit Leuven, Leuven, Belgium, in 1996 and 2002, respectively. After a Postdoctoral Fellowship of the Fund for Scientific Research, Flanders, Belgium, he cofounded Kimotion Technologies, Inc., Geneva, Switzerland. In 2006, he joined the University College “Karel de Grote,” Hoboken, where he was appointed as the Vice Dean of Academic Affairs. In 2013, he was one of the main founders of the Faculty of Applied Engineering, University of Antwerp, Antwerp, Belgium. Since then, he has been a Professor with the Faculty of Applied Engineering. Until 2023, he was the Chairperson of the Faculty’s Educational Board and the Vice Dean for Education of the Faculty. He is a member of staff of the Cosys-Laboratory, Antwerp. His research interests include signal and sensor processing and the implementation and automatic generation of these systems in embedded technology for applications in industry and (health) care.
\end{IEEEbiography}

\begin{IEEEbiography}[{\includegraphics[width=1in,height=1.25in,clip,keepaspectratio]{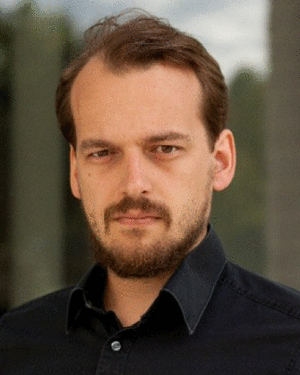}}]{Siegfried Mercelis}
received a master’s degree in Music Production from the Royal Conservatory of Ghent, Belgium, in 2008, a Master of Science in Applied Engineering (Electronics and ICT) from Karel de Grote University College in 2012, where he received the VIK Award for his master’s thesis on parallel data structures, and a Ph.D. degree in Applied Engineering from the University of Antwerp, Belgium, in 2016. From 2012 to 2016, he was with Van den Berghe R\&D under a Baekeland Ph.D. mandate, working on the optimization and parallelization of real-time media applications. He is currently an Assistant Professor with the IDLab research group at the University of Antwerp and imec, where he leads the Adaptive Intelligence research program, focusing on bridging academic AI research and industrial applications in domains such as process control, autonomous shipping, logistics, and mobility. He has authored over 100 peer-reviewed publications.
\end{IEEEbiography}

\begin{IEEEbiography}[{\includegraphics[width=1in,height=1.25in,clip,keepaspectratio]{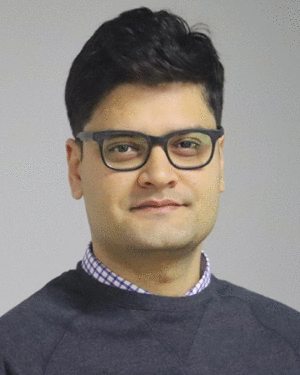}}]{Ali Anwar} (Member, IEEE) received his Ph.D. degree in control science and engineering from Harbin Institute of Technology, Harbin, China, in 2019. Since 2020, he has been a Principal Research Fellow with IDLab, an IMEC research group with the University of Antwerp, Belgium, where he currently leads a team on context-aware control systems. Since, 2024, he serves as a principal investigator for context-aware control systems at the Adapt-I program of IDLab. His research interests include autonomous vessel navigation, safe and robust reinforcement learning, cooperative perception, and generative modeling in computer vision. He is with the IEEE Industrial Electronics and Systems, Man and Cybernetics Society, where he is part of the technical committees on motion control, and control, robotics, and mechatronics. He is a reviewer in journals including IEEE TRANSACTIONS ON INDUSTRIAL ELECTRONICS, IEEE TRANSACTIONS ON INDUSTRIAL INFORMATICS, IEEE/ASME TRANSACTIONS ON MECHATRONICS, and IEEE ACCESS. Moreover, he also reviews for conferences such as CVPR, IECON, ICRA and IROS.
\end{IEEEbiography}

\vfill
\EOD
\end{document}